\documentclass[10pt,journal,compsoc]{IEEEtran}
\ifCLASSOPTIONcompsoc
  \usepackage[nocompress]{cite}
\else
  \usepackage{cite}
\fi
\ifCLASSINFOpdf
  \usepackage[pdftex]{graphicx}
\else
\fi
\usepackage{amsmath}
\usepackage{array}
\usepackage{url}

\usepackage{xcolor}
\usepackage{booktabs}
\usepackage{multirow}
\usepackage{amssymb}
\usepackage{dsfont}
\usepackage{bm}

\newcommand{\eg}{e.g.\ }
\newcommand{\etal}{et al.\ }
\newcommand{\ie}{i.e.\ }
\newcommand{\wrt}{w.r.t.\ }
\newcommand{\rgbd}{\mbox{RGB-D} }

\DeclareMathOperator*{\argmin}{\arg\!\min}
\DeclareMathOperator*{\argmax}{\arg\!\max}

\newcommand{\est}[1]{\hat{#1}}
\newcommand{\gt}[1]{{#1}^*}
\newcommand{\set}[1]{\mathcal{#1}}

\newcommand{\pos}{\mathbf{p}}
\newcommand{\eye}{\mathbf{e}}

\newcommand{\crd}{\mathbf{y}}
\newcommand{\crds}{\mathcal{Y}}
\newcommand{\refine}{\mathbf{R}}

\newcommand{\mdl}{\mathbf{h}}

\newcommand{\loss}{\ell}
\newcommand{\Loss}{\mathcal{L}}

\newcommand{\param}{\mathbf{w}}

\newcommand{\expectation}[2]{\mathbb{E}_{#1}\left[ #2 \right]}
\newcommand{\derv}[1]{\frac{\partial}{\partial #1}}

\begin{document}
%
\title{Visual Camera Re-Localization \\from RGB and \rgbd Images Using DSAC}
%
%
%
%

\author{Eric~Brachmann
        and~Carsten~Rother
\thanks{E.~Brachmann is with Niantic. Work done during his time at the Visual Learning Lab, Heidelberg University.}
\thanks{C.~Rother is with the Visual Learning Lab, Heidelberg University.\vspace{0.2cm}}
\thanks{This work has been submitted to the IEEE for possible publication. Copyright may be transferred without notice, after which this version may no longer be accessible.}
}

\IEEEtitleabstractindextext{%
\begin{abstract}
We describe a learning-based system that estimates the camera position and orientation from a single input image relative to a known environment.
The system is flexible \wrt the amount of information available at test and at training time, catering to different applications.
Input images can be \rgbd or RGB, and a 3D model of the environment can be utilized for training but is not necessary.
In the minimal case, our system requires only RGB images and ground truth poses at training time, and it requires only a single RGB image at test time. 
The framework consists of a deep neural network and fully differentiable pose optimization.
The neural network predicts so called scene coordinates, \ie dense correspondences between the input image and 3D scene space of the environment.
The pose optimization implements robust fitting of pose parameters using differentiable RANSAC (\emph{DSAC}) to facilitate end-to-end training.
The system, an extension of DSAC++ and referred to as DSAC*, achieves state-of-the-art accuracy an various public datasets for RGB-based re-localization, and competitive accuracy for \rgbd based re-localization.
\end{abstract}

\begin{IEEEkeywords}
Camera Re-Localization, Pose Estimation, Differentiable RANSAC, DSAC, Differentiable Argmax, Differentiable PnP
\end{IEEEkeywords}}

\maketitle

\IEEEdisplaynontitleabstractindextext

%
\IEEEpeerreviewmaketitle

\IEEEraisesectionheading{\section{Introduction}\label{sec:introduction}}

%
%
%
%
\IEEEPARstart{T}{he} ability to re-localize ourselves has revolutionized our daily lives. 
GPS-enabled smart phones already facilitate car navigation without a co-driver sweating over giant fold-out maps, or they enable the search for a rare vegetarian restaurant in the urban jungle of Seoul. 
On the other hand, the limits of GPS-based re-localization are clear to anyone getting lost in vast indoor spaces or in between sky scrapers.
When the satellite signals are blocked or delayed, GPS does not work or becomes inaccurate.
At the same time, upcoming technical marvels, like autonomous driving \cite{Geiger2012kitti} or impending updates of reality itself (\ie augmented/extended/virtual reality \cite{mann2018all}), call for reliable, high precision estimates of camera position and orientation.

Visual camera re-localization systems offer a viable alternative to GPS by matching an image of the current environment, \eg taken by a handheld device, with a database representation of said environment.
From a single image, state-of-the-art visual re-localization methods estimate the camera position to the centimeter, and the camera orientation up to a fraction of a degree, both indoors and outdoors. 

Existing re-localization approaches rely on varying types of information to solve the task, effectively catering to different application scenarios. 
Some use \rgbd images as input which facilitates highest precision suitable for augmented reality \cite{shotton13scorf,valentin2015cvpr,brachmann2016, schmidt2017ssvdl}.
However, they require capturing devices with active or passive stereo capabilities, where the former only works indoors, and the latter requires a large stereo baseline for reliable depth estimates outdoors.
Approaches based on feature-matching use an RGB image as input and also offer high precision \cite{sattler2016efficient}.
But they require a structure-from-motion (SfM) reconstruction \cite{Snavely2006bundler,visualsfm13,schoenberger2016sfm} of the environment for re-localization.
Such reconstructions might be cumbersome to obtain indoors due to texture-less surfaces and repeating structures obstructing reliable feature matching \cite{LSTMPoseNet}.
Finally, approaches based on image retrieval or pose regression require only a database of RGB images and ground truth poses for re-localization, but suffer from low precision comparable to GPS \cite{sattler2019limits}.

\begin{figure}[!t]
    \centering
    \vspace{-1.5cm}
    \includegraphics[width=1\linewidth]{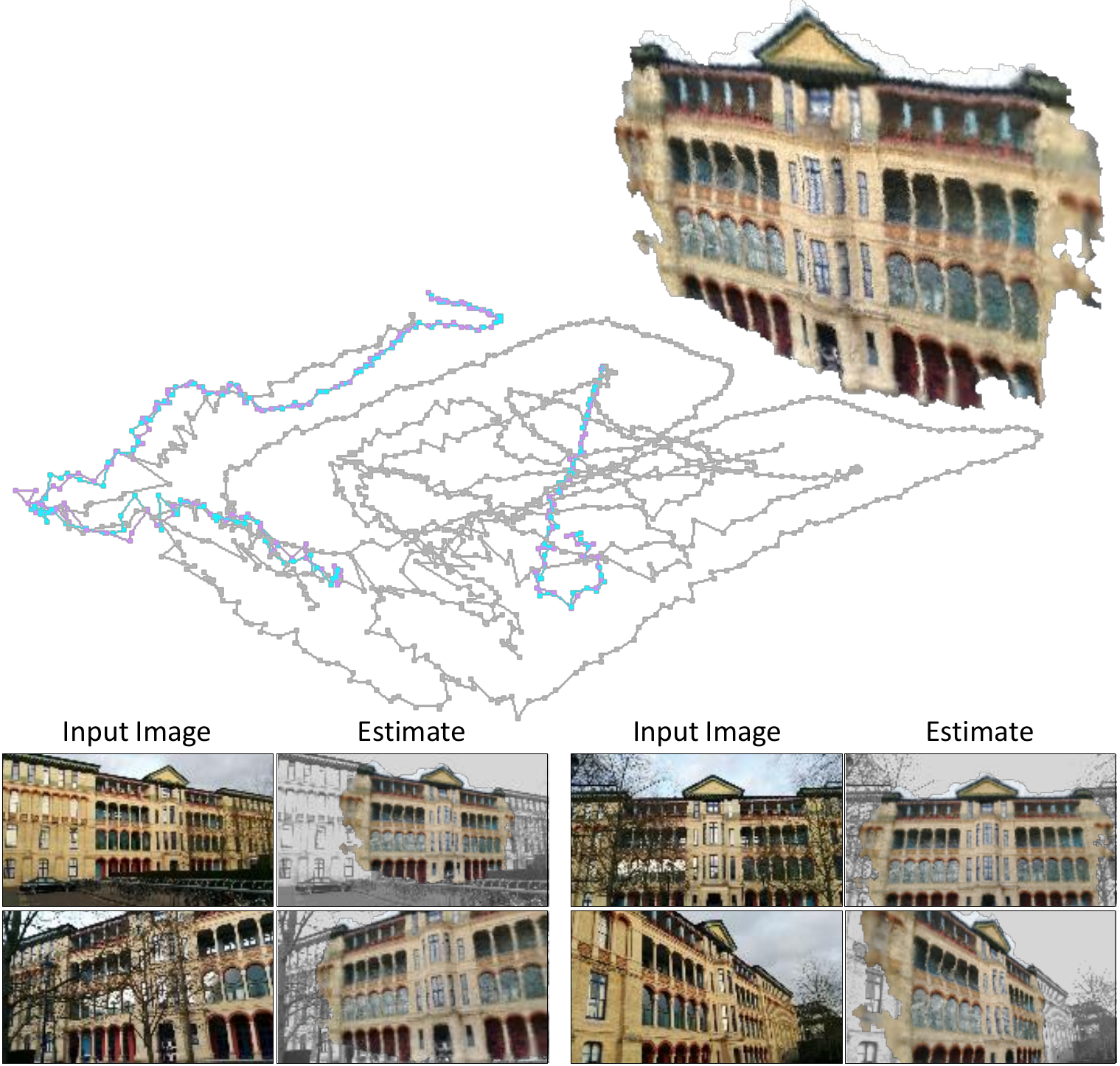}
    \vspace{-0.7cm}
    \caption{\textbf{Top.} Our system accurately re-localizes within a known environment given a single image. We show estimated camera positions in {\color{violet}purple} and ground truth in {\color{cyan}cyan}. In this instance, the system was trained using RGB images and associated ground truth poses, only ({\color{gray}gray} trajectory), In particular, the scene geometry, displayed as a 3D model, was discovered by the system, automatically. \textbf{Bottom.} To visualize the re-localization quality, we render the learned 3D geometry using estimated poses over gray-scale input images.}
    \vspace{-0.5cm}
    \label{fig:teaser}
\end{figure}

In this work, we describe a versatile, learning-based framework for visual camera re-localization that covers all aforementioned scenarios.
In the minimal case, it requires only a database of RGB images and ground truth poses of an environment for training, and re-localizes based on a single RGB image at test time with high precision.
In such a scenario the system automatically discovers the 3D geometry of the environment during training, see Fig.~\ref{fig:teaser} for an example.
If a 3D model of the scene exists, either as a SfM reconstruction or a 3D scan, we can utilize it to help the training process.
The framework exploits depth information at training or test time if an RGB-D sensor is available.

We base our approach on scene coordinate regression initially proposed by Shotton \etal \cite{shotton13scorf} for \mbox{RGB-D}-based camera re-localization.
A learnable function, a random forest in \cite{shotton13scorf}, regresses for each pixel of an input image the corresponding 3D coordinate in the environment's reference frame.
This induces a dense correspondence field between the image and the 3D scene that serves as basis for RANSAC-based pose optimization.
In our work, we replace the random forest of \cite{shotton13scorf} with a fully convolutional neural network \cite{fcn2015}, and derive differentiable approximations to all steps of pose optimization.
Most prominently, we derive a differentiable approximation of the RANSAC robust estimator, called \emph{differentiable sample consensus} (DSAC) \cite{brachmann2017dsac}.
Additionally, we describe an efficient differentiable approximation for calculating gradients of the perspective-n-point problem \cite{brachmann2018lessmore}.
These ingredients make our framework end-to-end trainable, ensuring that the neural network predicts scene coordinates that result in high precision camera poses.

This article is a summary and extension of our previous work on camera re-localization published in \cite{brachmann2017dsac} as DSAC, and its follow-up DSAC++ \cite{brachmann2018lessmore}.
In particular, we describe an improved version under the name \emph{DSAC*} with the following properties.
\begin{itemize}
\item We extent DSAC++ to optionally utilize \rgbd inputs. 
The corresponding pose solver is naturally differentiable, and other components require only minor adjustments. 
When using RGB-D, DSAC* achieves accuracy comparable to state-of-the-art accuracy on standard indoor re-localization datasets.
\item We propose a simplified training procedure which unifies the two separate initialization steps used in DSAC++.
As a result, the training time of DSAC* reduces from 6 days to 2.5 days on identical hardware. 
\item The improved initialization also leads to better accuracy. 
Particularly, when training without a 3D model, results improve significantly from 53.1\% (DSAC++) to 80.7\% (DSAC*) for indoor re-localization.
\item We utilize an improved network architecture for scene coordinate regression which we introduced in \cite{brachmann2019neural, brachmann19esac}.
The architecture, based on ResNet \cite{resnet2015}, reduces the memory footprint by 75\% compared to the network of DSAC++.
A forward pass of the new network takes 50ms instead of 150ms on identical hardware. 
Together with better pose optimization parameters we decrease the total inference time from 200ms for DSAC++ to 75ms for DSAC*.
\item In new ablation studies, we investigate the impact of training data augmentation, the impact of the  network's receptive field, as well as the impact of end-to-end training. We also analyze the scene compression properties of DSAC*.
Furthermore, we provide extensive visualizations of our pose estimates, and of the 3D geometry that the network encodes.
\item We migrate our implementation of DSAC++ from LUA/Torch to PyTorch \cite{paszke2017automatic} and make it publicly available: \url{https://github.com/vislearn/dsacstar}
\end{itemize}

This article is organized as follows: We give an overview of related work in Sec.~\ref{sec:related}. 
In Sec.~\ref{sec:framework}, we formally introduce the task of camera re-localization and how we solve it via scene coordinate regression.
In Sec.~\ref{sec:score}, we discuss how to train the scene coordinate network using auxiliary losses defined on the scene coordinate output.
In Sec.~\ref{sec:diffpose}, we discuss how to train the whole system end-to-end, optimizing a loss on the estimated camera pose.
We present experiments for indoor and outdoor camera re-localization, including ablation studies in Sec.~\ref{sec:experiments}.
We conclude this article in Sec.~\ref{sec:conclusion}.

\section{Related Work}
\label{sec:related}
In the following, we discuss the main strains of research for solving visual camera re-localization. 
We also discuss related work on differentiable robust estimators other than DSAC.

\subsection{Image Retrieval and Pose Regression}

Early examples of visual re-localization rely on efficient image retrieval \cite{schindler2007city}.
The environment is represented as a collection of data base images with known camera poses.
Given a query image, we search for the most similar data base image by matching global image descriptors, such as DenseVLAD\cite{torii2015viewsynth}, or its learned successor NetVLAD \cite{netvlad2016}.
The metric to compare global descriptors can be learned as well \cite{cao2013graph}.
The sampling density of data base images inherently limits the accuracy of retrieval-based system.
However, they scale to very large environments, and can serve as an efficient initialization for local pose refinement \cite{sattler2017largescale, taira2018inloc}.

Absolute pose regression methods \cite{kendall2015convolutional,LSTMPoseNet,geometricloss,naseer2017svspose,mapnet2018} aim at overcoming the precision limitation of image retrieval while preserving efficiency and scalability.
Interpreting the data base images as a training set, a neural network learns the relationship between image content and camera pose.
In theory, the network could learn to interpolate poses of training images, or even generalize to novel view points. 
In practise, however, absolute pose regression fails to consistently outperform the accuracy of image retrieval methods \cite{sattler2019limits}.

Relative pose regression methods \cite{relocnet2018,saha2018anchornet} train a neural network to predict the relative transformation between the query image, and the most similar data base image found by image retrieval.
Initial relative pose regression methods suffered from similarly low accuracy as absolute pose regression \cite{sattler2019limits}.
However, recent work \cite{ding2019camnet} suggests that relative pose regression can achieve accuracy comparable to structure-based methods which we discuss next. 

\subsection{Sparse Feature Matching}

The camera pose can be recovered by matching sparse, local features like SIFT \cite{Lowesift} between the query image and database images \cite{Zhang2006urban}.
For an efficient data base representation, SfM tools \cite{visualsfm13,schoenberger2016sfm} create a sparse 3D point cloud of an environment, where each 3D point has one or several feature descriptors attached to it.
Given a query image, feature matching established 2D-3D correspondences which can be utilized in RANSAC-based pose optimization to yield a very precise camera pose estimate \cite{sattler2016efficient}.
Work on feature-based re-localization has primarily focused on scaling to very large environments \cite{li2012worldwide,svarm2014accurate,sattler2015hyperpoints,sattler2016large,sattler2017largescale,svarm2017city} enabling city or even world scale re-localization.
Other authors worked on efficiency to run feature-based re-localization on mobile devices with low computational budget \cite{lim2012real}.

While sparse feature matching can achieve high re-localization accuracy, hand-crafted features fail in certain scenarios.
Feature detectors have difficulty finding stable points under motion blur \cite{kendall2015convolutional} and for texture-less areas \cite{shotton13scorf}. 
Also, SfM reconstructions tend to fail in indoor environments dominated by ambiguous, repeating structures \cite{LSTMPoseNet}.
Learning-based sparse feature pipelines \cite{lift2016,detone18superpoint,Revaud2019r2d2,Dusmanu2019d2net} might ultimately be able to overcome these issues, but currently it is an open research question whether learned sparse features consistently exceed the capabilities of their hand-crafted predecessors \cite{schonberger2017comparative,bhowmik2019reinforced}. 

State-of-the-art feature-based re-localization methods such as ActiveSearch \cite{sattler2016efficient} offer no direct possibility to incorporate depth sensors when available at test time, neither do current state-of-the-art SfM tools like COLMAP \cite{schoenberger2016sfm} support depth sensors when creating the scene reconstruction.

\subsection{Scene Coordinate Regression}

Instead of relying on a feature detector to identify salient image structures suitable for discrete matching,
scene coordinate regression \cite{shotton13scorf} predicts the corresponding 3D scene point for a given 2D pixel location, directly. 
In these works, the environment is implicitly represented by a learnable function that can be evaluated for any image pixel to predict a dense correspondence field between image and scene.
The correspondences serve as input for RANSAC-based pose optimization, similar to sparse feature techniques.

Originally, scene coordinate regression was proposed for \mbox{RGB-D}-based re-localization in indoor environments \cite{shotton13scorf, guzman2014multi, valentin2015cvpr, meng18}.
The depth channel would serve as additional input to a scene coordinate regression forest, and be used in pose optimization by allowing to establish and resolve 3D-3D correspondences \cite{kabsch1976solution}.
Scene coordinate regression forests were later shown to also work well for RGB-based re-localization \cite{brachmann2016, meng17}.

Recent works on scene coordinate regression often replace the random forest regressor by a neural network while continuing to focus on RGB inputs \cite{rfvscnn2016, brachmann2017dsac, brachmann2018lessmore, li2018abrloss, brachmann19esac}.
In previous work, we have shown that the RANSAC-based pose optimization can be made differentiable to allow for end-to-end training of a scene coordinate regression pipeline \cite{brachmann2017dsac, brachmann2018lessmore}.
In particular, \cite{brachmann2017dsac} introduced a differentiable approximation of RANSAC \cite{ransac1981}, and \cite{brachmann2018lessmore} described an efficient analytical approximation of calculating gradients for perspective-n-point solvers.
Furthermore, the predecessor of the current work, DSAC++ \cite{brachmann2018lessmore}, introduced the possibility to train scene coordinate regression solely from RGB images and ground truth poses, without the need for image depth or a 3D model of the scene.
Li \etal \cite{li2018abrloss} improved on this initial effort by enforcing multi-view and photometric consistency throughout training.
In a follow-up work, Li \etal \cite{li2020hierarchical} introduce a joint classification-regression network architecture for predicting scene coordinate, and demonstrate the effectiveness of training data augmentation for large improvements on standard benchmarks.

In this work, we describe several improvements to DSAC++ that increase accuracy while reducing training and test time. 
We demonstrate that the DSAC framework naturally exploits image depth if available, in an attempt to unify previously distinct strains of RGB- and RGB-D-based re-localization research.
In summary, our method is more precise and more flexible than previous scene coordinate regression- and sparse feature-based re-localization systems.
At the same time, it is as simple to deploy as absolute pose regression systems due to requiring only a set of RGB images with ground truth poses for training in the minimal setting.

Orthogonal to this work, we describe a scalable variant of DSAC-based re-localization in \cite{brachmann19esac}.
Yang \etal explore the possibility of allowing for scene-independent coordinate regression \cite{yang2019sanet}, and Cavallari \etal  adapt scene coordinate regression forests and networks on-the-fly for deployment as a re-localizer in simultaneous localization and mapping (SLAM) \cite{cavallari2017fly, Cavallari2019cascade,Cavallari2019flynet}.

\subsection{Differentiable Robust Estimators}

To allow for end-to-end training of our re-localization pipeline, we have introduced a differentiable approximation to the RANSAC \cite{ransac1981} algorithm, called \emph{differentiable sample consensus} (DSAC).
DSAC relies on a formulation of RANSAC that reduces to a $\argmax$ operation over model parameters.
Instead of choosing model parameters with maximum consensus, we choose model parameters randomly with a probability \emph{proportional to consensus}.
This allows us to optimize the expected task loss for end-to-end training. 
A DSAC variant using a soft $\argmax$ \cite{Chapelle2010} does not work as well since it ignores potential multi-modality in the distribution of model parameters.
Recently, Lee \etal proposed a kernel soft $\argmax$ as an alternative that is robust to multiple modes in the arguments \cite{lee2019sfnet}. 
However, their approximation effectively suppresses gradients of all but the main mode, while the DSAC estimator utilizes gradients of all modes.

Alternatively to making RANSAC differentiable, some authors propose to replace RANSAC by a neural network \cite{goodcorr18,Zhang2019orderaware,deepfund18,rocco2018connet,probst2019unsmax}. 
In these works, the neural network acts as a classifier for model inliers, effectively acting as a robust estimator for model parameters.
However, \emph{NG-RANSAC} \cite{brachmann2019neural} demonstrates that the \emph{combination} of an inlier-scoring network and RANSAC achieves even higher accuracy. 
In \cite{brachmann2019neural}, we also discuss a combination of NG-RANSAC and DSAC for camera re-localization which leads to higher accuracy in outdoor re-localization by learning to focus on informative image areas.

\section{Framework}
\label{sec:framework}

In this section, we introduce the task of camera re-localization, and the principle of scene coordinate regression \cite{shotton13scorf}.
We explain how to estimate the camera pose from scene coordinates using RANSAC \cite{ransac1981} when the input is a single RGB or \rgbd image, respectively.

\begin{figure}[!t]
    \centering
    \includegraphics[width=1\linewidth]{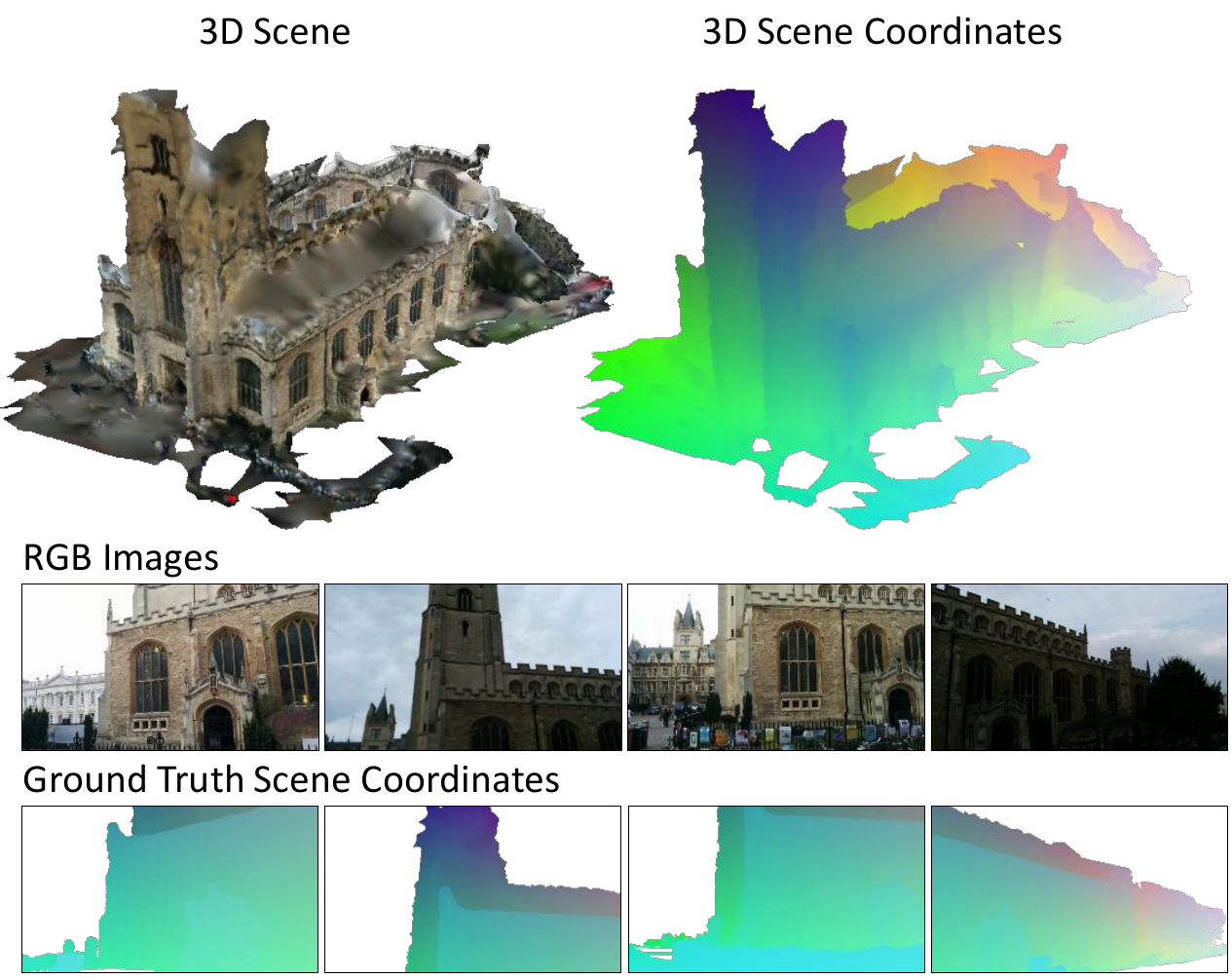}
    \vspace{-0.5cm}
    \caption{\textbf{Scene Coordinates \cite{shotton13scorf}. Top.} Every surface point in a 3D environment has a unique 3D coordinate in the local coordinate frame. We visualize 3D scene coordinates by mapping XYZ to the RGB cube. \textbf{Bottom.} A 3D scene model together with ground truth camera poses allows us to render ground truth scene coordinates for images, \eg to serve as training targets. We can also create training targets from depth maps instead of a 3D model, or from ground truth poses alone by optimizing the re-projection error over multiple frames.}
    \label{fig:method:scenecoordinates}
\end{figure}

\begin{figure*}[!t]
    \centering
    \includegraphics[width=1\linewidth]{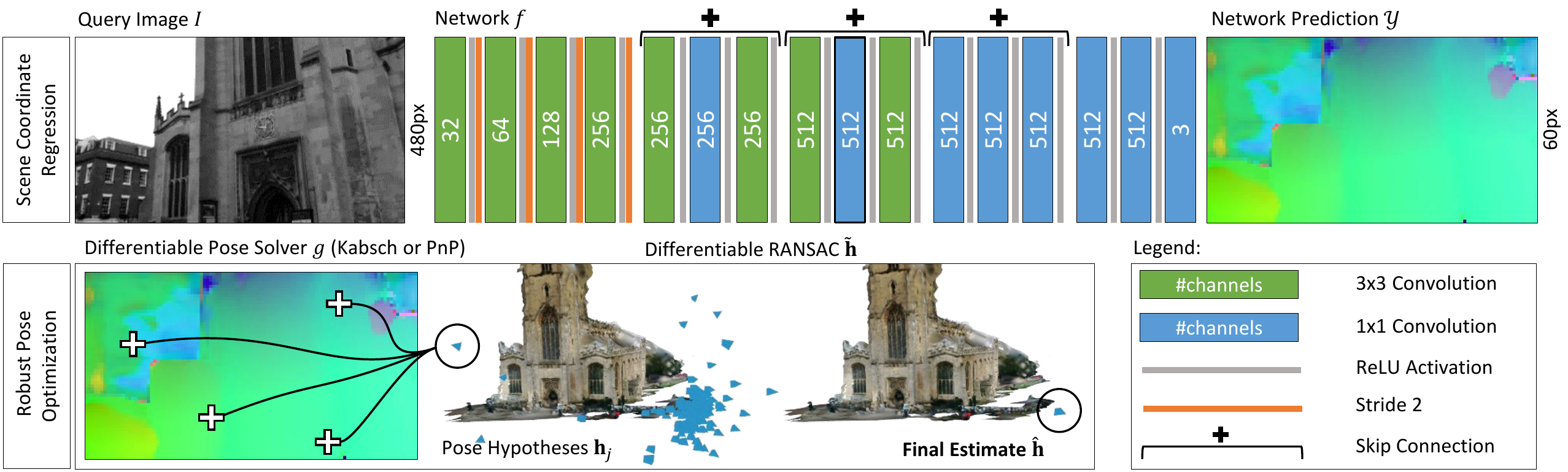}
    \caption{\textbf{System Overview.} The system consists of two stages: Scene coordinate regression using a CNN (\textbf{top}) and differentiable pose estimation (\textbf{bottom}). The network is fully convolutional and produces a dense but sub-sampled output. Pose estimation employs a minimal solver (PnP\cite{gao2003complete} for RGB images or Kabsch\cite{kabsch1976solution} for \rgbd images) within a RANSAC\cite{ransac1981} robust estimator. The final camera pose estimate is also refined. To allow for end-to-end training, all components need to be differentiable. While the Kabsch solver is inherently differentiable, we describe differentiable approximations for PnP and RANSAC.} 
    \label{fig:method:system}
\end{figure*}

Given an image $I$, which can be either RGB or \mbox{RGB-D}, we aim at estimating camera pose parameters $\mdl$ \wrt the reference coordinate frame of a known scene, a task  called re-localization.
We propose a learnable system to solve the task, which is trained for a specific scene to re-localize within that scene.
The camera pose has 6 degrees of freedom (DoF) corresponding to the 3D camera position $\mathbf{t}$ and its 3D orientation $\boldsymbol{\theta}$.
In particular, we define the camera pose as the transformation that maps 3D points in the camera coordinate space, denoted as $\eye$ to 3D points in scene coordinate space, denoted as $\crd$, \ie 
\begin{equation}
    \label{eq:scenecoordinates}
    \crd_i = \mdl \eye_i,
\end{equation}
where $i$ denotes the pixel index in image $I$.
For notational simplicity, we assume a 4x4 matrix representation of the camera pose $\mdl$ and homogeneous coordinates for all points where convenient.  

We denote the complete set of scene coordinates for a given image as $\crds$, \ie $\crd_i \in \crds$. 
See Fig.~\ref{fig:method:scenecoordinates} for an explanation and visualization of scene coordinates.
Originally proposed by Shotton \etal \cite{shotton13scorf}, scene coordinates $\crds$ induce a dense correspondence field between camera coordinate space and scene coordinate space which we can use to solve for the camera pose.
To estimate $\crds$ for a given image, we utilize a neural network $f$ with learnable parameters $\param$:
\begin{equation}
    \crds = f(I; \param).
\end{equation}
Due to potential errors in the neural network prediction, we utilize a robust estimator, namely RANSAC \cite{ransac1981}, to recover $\mdl$ from $\crds$.
Our RANSAC-based pose optimization consists of the following steps:
\begin{enumerate}
    \item Sample a set of camera pose hypotheses.
    \item Score each hypothesis and choose the best one.
    \item Refine the winning hypothesis.
\end{enumerate}
We show an overview of our system in Fig.~\ref{fig:method:system}.
In the following, we describe the three aforementioned steps for the general case, while we elaborate on concrete manifestations for RGB and \rgbd input images in Sec.~\ref{sec:framework:rgb} and Sec.~\ref{sec:framework:rgbd}, respectively. 

\noindent \textbf{1) Sample Hypotheses.}
Image $I$ and scene coordinate prediction $\crds$ define a dense correspondence field $\set{C}$ over all image pixels $i$.
We will specify the concrete nature of correspondences in sub-sections below because it differs for RGB and \rgbd inputs.
As first step of robust pose optimization we randomly choose $M$ subsets of correspondences, $\set{C}_j \subseteq \set{C}$, with $0 \leq j < M$.
Each correspondence subset  $\set{C}_j$ corresponds to a camera pose hypothesis $\mdl_j$, which we recover using a pose solver $g$, \ie
\begin{equation}
\mdl_j = g(\set{C}_j).
\end{equation}
The concrete manifestation of $g(\cdot)$ differs for RGB and \rgbd inputs. 
Note that the RANSAC algorithm \cite{ransac1981} includes a way to adaptively choose the number of hypotheses $M$ according to an online estimate of the outlier ratio in $\set{C}$, \ie the amount of erroneous correspondences.
In this work, and our previous work \cite{brachmann2016, brachmann2017dsac, brachmann2018lessmore, brachmann19esac, brachmann2019neural}, we choose a fixed $M$ and train the system to adapt to this particular setting.
Thereby, $M$ becomes a hyper-parameter that controls the allowance of the neural network $f$ to make inaccurate predictions. 

\noindent \textbf{2) Choose Best Hypothesis.}
Following RANSAC, we choose the hypothesis $\mdl_j$ with maximum consensus among all scene coordinates $\crds$, \ie
\begin{equation}
    \label{eq:framework:ransac}
    \tilde{\mdl} = \argmax_{\mdl_j} s(\mdl_j, \crds).
\end{equation}
We measure consensus by a scoring function $s(\cdot)$ that is, by default, implemented as inlier counting:
\begin{equation}
    \label{eq:hardinliers}
    s(\mdl, \crds) = \sum_{\crd_i \in \crds}  \mathds{1}[r(\crd_i, \mdl) < \tau].
\end{equation}
Function $r(\cdot)$ measures the residual between pose parameters $\mdl$, and a scene coordinate $\crd_i$, $\mathds{1}[\cdot]$ evaluates to one if the residual is smaller than an inlier threshold $\tau$. 

\noindent \textbf{3) Refine Best Hypothesis.}
We refine the chosen hypothesis $\tilde{\mdl}$, which was created from a small subset of correspondences, using all scene coordinates:
\begin{equation}
    \label{eq:framework:refine}
    \est{\mdl} = \refine(\tilde{\mdl}, \crds).
\end{equation}
We implement refinement as re-solving for the pose parameters using the complete inlier set $\set{I}$ of hypothesis $\tilde{\mdl}$, \ie
\begin{equation}
    \refine(\tilde{\mdl}, \crds) = g(\set{C}_\set{I}) \text{ with } \set{I} = \{i|r(\crd_i, \tilde{\mdl}) < \tau \}
\end{equation}
In practise, we iterate refinement and re-calculation of the inlier set $\set{I}$ until convergence.
We refer to the refined, chosen hypothesis as our final camera pose estimate $\est{\mdl}$.

Next, we discuss particular choices for pose optimization components in case the input image is RGB or \rgbd.

\subsection{Case RGB}
\label{sec:framework:rgb}

In case the input is an RGB image without a depth channel, correspondences $\set{C}$ manifest as 2D-3D correspondences between the image and 3D scene space:
\begin{equation}
\set{C}^\text{RGB} = \{(\pos_i, \crd_i) | \crd_i \in \crds\},
\end{equation}
where $\pos_i$ denotes the 2D image coordinate associated with pixel $i$.
Image coordinates and scene coordinates are related by 
\begin{equation}
\pos_i = K\mdl^{-1}\crd_i    
\end{equation}
where $K$ denotes the camera calibration matrix, or internal calibration parameters of the camera.
Using this relation, perspective-n-point (PnP) solvers $g(\cdot) $\cite{gao2003complete, mutliview2004} recover the camera pose from at least four 2D-3D correspondences: $|\set{C}_j^\text{RGB}| \ge 4$.
In practise, we use $|\set{C}_j^\text{RGB}| = 4$ with the solver of Gao \etal \cite{gao2003complete} when sampling pose hypotheses $\mdl_j$, and non-linear optimization of the re-projection error with Levenberg-Marquardt \cite{Levenberg44lm, Marquardt63lm} when refining the chosen hypothesis $\refine(\tilde{\mdl}, \crds)$ with $|\set{C}_\set{I}| > 4$.
We utilize the implementation available in OpenCV \cite{opencv_library} for all PnP solvers.

As residual function $r(\cdot)$ for determining the score $s(\cdot)$ of a pose hypothesis $\mdl_j$ in Eq.~\ref{eq:hardinliers}, we calculate the re-projection error:
\begin{equation}
\label{eq:residual:rgb}
    r^\text{RGB}(\crd_i, \mdl) = ||\pos_i - K\mdl^{-1}\crd_i||.
\end{equation}

\subsection{Case RGB-D}
\label{sec:framework:rgbd}

In case the input is an \rgbd image, the known depth map allows us to recover the 3D coordinate corresponding to each pixel $i$ in the coordinate frame of the camera, denoted as $\eye_i$.
Together with the scene coordinate prediction $\crds$, we have dense 3D-3D correspondences $\set{C}$ between camera space and scene space, \ie
\begin{equation}
\set{C}^\text{RGB-D} = \{(\eye_i, \crd_i) | \crd_i \in \crds\}.
\end{equation}
To recover the camera pose from 3D-3D correspondences we utilize the Kabsch algorithm \cite{kabsch1976solution}, sometimes also called orthogonal Procrustes, as pose solver $g(\cdot)$.
For sampling pose hypotheses $\mdl_j$, we use $|\set{C}_j^\text{RGB-D}| = 3$, when refining the chosen hypothesis $\refine(\tilde{\mdl}, \crds)$ we use $|\set{C}_\set{I}| > 3$.

As residual function $r(\cdot)$ for determining the score $s(\cdot)$ of an hypothesis $\mdl_j$ in Eq.~\ref{eq:hardinliers}, we calculate the 3D Euclidean distance:
\begin{equation}
    r^\text{RGB-D}(\crd_i, \mdl) = ||\eye_i - \mdl^{-1}\crd_i||.
\end{equation}

\section{Deep Scene Coordinate Regression}
\label{sec:score}

In this section, we discuss the neural network architecture for scene coordinate regression, and how to train it using auxiliary losses defined on the scene coordinate output. 
These auxiliary losses serve as an initialization step prior to training the whole pipeline in an end-to-end fashion, see Sec.~\ref{sec:diffpose}. 
The initialization is necessary, since end-to-end training from scratch will converge to a local minimum without giving reasonable pose estimates.

We implement scene coordinate regression $f(\cdot)$ using a fully convolutional neural network \cite{fcn2015} with skip connections \cite{resnet2015} and learnable parameters $\param$.
We depict the network architecture in Fig.~\ref{fig:method:system}, top.
The network takes a single channel grayscale image as input, and produces a dense scene coordinate prediction sub-sampled by the factor 8.
Sub-sampling, implemented with stride 2 convolutions, increases the receptive field associated with each pixel output while also enhancing efficiency.
The total receptive field of each output scene coordinate is 81px.
In experiments on various datasets, we found no advantage in providing the full RGB image as input, in contrast, conversion to grayscale slightly increases the robustness to non-linear lighting effects.

\noindent{\textbf{Relation to our Previous Work.}}
In our first DSAC-based re-localization pipeline \cite{brachmann2017dsac} and in DSAC++ \cite{brachmann2018lessmore}, we utilized a VGGNet-style architecture \cite{Simonyan2014vgg}.
It had a larger memory footprint and slower runtime while offering similar accuracy. 
The receptive field was comparable with 79px.
In the experiments of Sec.~\ref{sec:experiments}, we conduct an empirical comparison of both architectures.
We utilized our updated architecture already in our work on ESAC \cite{brachmann19esac} and NG-RANSAC \cite{brachmann2019neural}.

\begin{table}[]
\caption{\textbf{Information Available at Training and Test Time.} ``D'' stands for depth channel, ``poses'' stands for ground truth camera poses. The 3D model of the scene may be a sparse point cloud, \eg from a SfM reconstruction \cite{schoenberger2016sfm, visualsfm13}, or a dense 3D scan \cite{izadi2011kinectfusion, newcombe2011kinectfusion, dai2017bundlefusion}.}
\centering
\begin{tabular}{l||cccc||cc}
 & \multicolumn{4}{c||}{Training} & \multicolumn{2}{c}{Test} \\ \hline
Setting & RGB & D & poses & 3D model & RGB & D \\ \hline
RGB-D & \checkmark & \checkmark & \checkmark &  & \checkmark & \checkmark \\
RGB + 3D model & \checkmark &  & \checkmark & \checkmark & \checkmark &  \\
RGB & \checkmark &  & \checkmark &  & \checkmark & 
\end{tabular}
\label{tab:settings}
\end{table}

In the following, we discuss different strategies on initializing the scene coordinate neural network, depending on what information is available for training.
In particular, we discuss training from \rgbd images for \mbox{RGB-D}-based re-localization, training from RGB images and a 3D model of the scene for RGB-based re-localization as well as training from RGB images only for RGB-based re-localization.
See Table \ref{tab:settings} for a schematic overview.
Other combinations are of course possible, \eg training from RGB images only, but having \rgbd images at test time.
However, we but restrict our discussion and experiments to the most common settings found in the literature \cite{shotton13scorf, brachmann2016, brachmann2018lessmore}.

\subsection{\rgbd}
\label{sec:score:rgbd}

For \mbox{RGB-D}-based pose estimation, we initialize our neural network by minimizing the Euclidean distance between predicted scene coordinates $\crd_i$ and ground truth scene coordinates $\gt{\crd}_i$.
\begin{equation}
\label{eq:loss:rgbd}
\loss^\text{RGB-D}(\crd_i,\gt{\crd}_i) = ||\gt{\crd}_i - \crd_i||
\end{equation}
We obtain ground truth scene coordinates $\gt{\crd}_i$ by re-projecting depth channels of training images to obtain 3D points $\eye_i$ in the camera coordinate frame, and transforming them using the ground truth pose $\gt{\mdl}$, \ie $\gt{\crd}_i = \gt{\mdl}\eye_i$.
We train the network using the average loss over all pixels of a training image:
\begin{equation}
\label{eq:avgloss:rgbd}
\Loss^\text{RGB-D}(\crds, \gt{\crds}) = \frac{1}{|\crds|} \sum_{\crd_i \in \crds} \loss^\text{RGB-D}(\crd_i, \gt{\crd}_i).
\end{equation}
We motivate optimizing the Euclidean distance for \mbox{RGB-D}-based re-localization by the fact that the corresponding Kabsch pose solver optimizes the pose over squared Euclidean residuals between camera coordinates and scene coordinates.
We found the plain, instead of the squared, Euclidean distance in Eq.~\ref{eq:loss:rgbd} superior in \cite{brachmann2017dsac} due to its robustness to outliers. 

\subsection{RGB + 3D Model}
\label{sec:score:rgbm}

In case the camera pose is to be estimated from an RGB image, the optimization of scene coordinates \wrt a 3D Euclidean distance is not optimal.
The PnP solver, which we utilize for pose sampling and pose refinement, optimizes the camera pose \wrt the re-projection error of scene coordinates.
Hence, for RGB-based pose estimation, we initialize the scene coordinate regression network by minimizing the re-projection error of its predictions, \ie $r^\text{RGB}(\crd_i, \gt{\mdl})$ where $r^\text{RGB}(\cdot)$ denotes the residual function defined for RGB in Eq.~\ref{eq:residual:rgb}, and $\gt{\mdl}$ denotes the ground truth camera pose.

Unfortunately, optimizing this objective from scratch fails since the re-projection error is ambiguous \wrt the viewing direction of the camera.
However, if we assume a 3D model of the environment to be available, we may render ground truth scene coordinates $\gt{\crds}$, optimize the \rgbd objective of Eq.~\ref{eq:loss:rgbd} first, and switch to the re-projection error after a few training iterations:
\begin{equation}
\label{eq:loss:rgbm}
    \loss^\text{RGB+M}(\crd_i, \gt{\crd}_i, \gt{\mdl})= 
\begin{cases}
     \est{r}^\text{RGB}(\crd_i, \gt{\mdl})  & \text{if } \crd_i \in \set{V}^\text{RGB+M}\\
    ||\gt{\crd}_i - \crd_i||                & \text{otherwise}.
\end{cases}
\end{equation}
We define a set of valid scene coordinate predictions as $\set{V}^\text{RGB+M}$ for which we optimize the re-projection error.
If a scene coordinate does not qualify as valid yet, we optimize the Euclidean distance, instead.
A prediction $\crd_i$ is valid, $\crd_i \in \set{V}^\text{RGB+M}$ iff:
\begin{enumerate}
    \item $({\gt{\mdl}}^{-1}\crd_i)_z > 0.1\text{m}$, \ie it lies at least 0.1m in front of the ground truth image plane.
    \item It has a maximum re-projection error of $r^\text{RGB}(\crd_i, \gt{\mdl}) < 1000\text{px}$.
    \item It is within a maximum 3D distance \wrt to the rendered ground truth coordinate of $||\gt{\crd}_i - \crd_i|| < 0.1\text{m}$.
\end{enumerate}
The training objective is flexible \wrt to missing ground truth scene coordinates for certain pixels, \ie if $\gt{\crd}_i = \mathbf{0}$.
In this case, we only enforce constraint 1) and 2) for $\set{V}^\text{RGB+M}$.
This allows us to utilize dense 3D models of the scene, sparse SfM reconstructions as well as depth channels with missing measurements to generate $\gt{\crd}_i$.
The training objective utilizes a robust version $\est{r}^\text{RGB}(\crd_i, \gt{\mdl})$ of the RGB residual function of Eq.~\ref{eq:residual:rgb}, \ie
\begin{equation}
\label{eq:loss:sqrtrgb}
\est{r}^\text{RGB}(\crd, \mdl) = \begin{cases}
    r^\text{RGB}(\crd, \mdl)                                        &\text{if }r^\text{RGB}(\crd, \mdl) < 100\text{px}\\
    \sqrt{100 r^\text{RGB}(\crd, \mdl)}    &\text{otherwise}.
    \end{cases}
\end{equation}
This formulation implements a soft clamping by using the square root of the re-projection residual after a threshold of 100px.
To train the scene coordinate network, we optimize the average of Eq.~\ref{eq:loss:rgbm} over all pixels of a training image, similar to Eq.~\ref{eq:avgloss:rgbd}.

\noindent{\textbf{Relation to our Previous Work.}}
We introduced a combined training objective based on, firstly, minimizing the 3D distance to ground truth scene coordinates, and, secondly, minimizing the re-projection error in DSAC++ \cite{brachmann2018lessmore}.
However, DSAC++ uses separate initalization stages for the two objectives, 3D distance and re-projection error, which is computationally wasteful.
The network might concentrate on modelling fine details of the geometry in the first initialization stage which is potentially undone in the second initialization stage.
Also, pixels without a ground truth scene coordinate would receive no training signal in the first initalization stage of DSAC++.
The new, combined training objective of DSAC* in Eq.~\ref{eq:loss:rgbm} switches dynamically from optimizing the 3D distance to optimizing the re-projection error on a per-pixel basis.
By using one combined initialization stage instead of two, we shorten the pre-training time of DSAC* from 4 days to 2 days compared to DSAC++ on identical hardware.

\subsection{RGB}

The previous RGB-based training objective of Eq.~\ref{eq:loss:rgbm} relies on the availability of a 3D model of the scene.
When a dense 3D scan of an environment is unavailable, SfM tools like VisualSfM \cite{visualsfm13} or COLMAP \cite{schoenberger2016sfm} offer workable solutions to create a (sparse) 3D model from a collection of RGB images, \eg from the training set of a scene.
However, for some environments, particularly indoors, a SfM reconstruction might fail due to texture-less areas or repeating structures.
Also, despite SfM tools having matured significantly over many years since the introduction of Bundler \cite{Snavely2006bundler} they still represent expert tools with their own set of hyper-parameters to be tuned.
Therefore, it might be attractive to train a camera re-localization system from RGB images and ground truth poses alone, without resorting to an SfM tool for pre-processing. 
Therefore, we introduce a variation on the RGB-based training objective of Eq.~\ref{eq:loss:rgbm} that substitutes ground truth scene coordinates $\gt{\crd}_i$ with a heuristic scene coordinate target $\bar{\crd}_i$ combined with a robust $L_1$ distance: 
\begin{equation}
\label{eq:loss:rgb}
    \loss^\text{RGB}(\crd_i, \bar{\crd}_i, \gt{\mdl})= 
\begin{cases}
    \est{r}^\text{RGB}(\crd_i, \gt{\mdl})   & \text{if } \crd_i \in \set{V}^\text{RGB}\\
    |\bar{\crd}_i - \crd_i|               & \text{otherwise}.
\end{cases}
\end{equation}
We obtain heuristic targets $\bar{\crd}_i = \gt{\mdl}\bar{\eye}_i$ from the ground truth camera pose $\gt{\mdl}$ and hallucinated 3D camera coordinates $\bar{\eye}_i$ re-projected by assuming a constant image depth of 10m.
The above formulation relies on switching from the heuristic target to the re-projection error as soon as possible.
Therefore, we formulate the following relaxed validity constraints for scene coordinate predictions $\crd_i$ to form the set $\set{V}^\text{RGB}$:
\begin{enumerate}
    \item $({\gt{\mdl}}^{-1}\crd_i)_z > 0.1\text{m}$, \ie it lies at least 0.1m in front of the ground truth image plane.
    \item $({\gt{\mdl}}^{-1}\crd_i)_z < 1000\text{m}$, \ie it lies at most 1000m in front of the ground truth image plane.
    \item It has a maximum re-projection error of $r^\text{RGB}(\crd_i, \gt{\mdl}) < 1000\text{px}$.
\end{enumerate}

\noindent{\textbf{Relation to our Previous Work.}}
DSAC++ \cite{brachmann2018lessmore} used two separate initialization stages for minimizing the distance to heuristic targets $\bar{\crd}_i$, and optimization of the re-projection error, respectively.
The first initialization stage was particularly cumbersome since the heuristic targets $\bar{\crd}_i$ are inconsistent \wrt the true 3D geometry of the scene.
The neural network can easily overfit to $\bar{\crd}_i$ which we circumvent in DSAC++ by early stopping and by using only a fraction of the full training data for the first initialization stage. 
The new, combined formulation of Eq.~\ref{eq:loss:rgb} is more robust by only loosely enforcing the heuristic until the formulation adaptively switches to the re-projection error.
Also, as mentioned in the previous section, the new formulation is more efficient by combining two initialization stages into one, thus reducing training time.

\section{Differentiable Pose Optimization}
\label{sec:diffpose}

Our overall goal is training the complete pose estimation pipeline in an end-to-end fashion.
That is, we wish to optimize the learnable parameters $\param$ of scene coordinate prediction in a way that we obtain highly accuracy pose estimates $\est{\mdl}$ as per Eq.~\ref{eq:framework:refine} and Eq.~\ref{eq:framework:ransac}.
Due to the robust nature of our pose optimization, particularly due to deploying RANSAC to estimate model parameters, the relation of the quality of scene coordinates $\crds$ and the estimated pose $\est{\mdl}$ is non-trivial.
For example, some predictions $\crd_i$ will be removed by RANSAC as outliers, hence they have no influence on $\est{\mdl}$. 
We may neglect such outlier scene coordinates entirely in training without any deterioration in accuracy.
To the contrary, it might be beneficial to \emph{decrease} the accuracy of outlier scene coordinates further to make sure that RANSAC classifies them as outliers.
However, we have no prior knowledge which exact predictions for an image should be inliers or outliers of the estimated model.

In this work, we address this problem by making pose optimization itself differentiable, to include it in the training process.
By training in an end-to-end fashion, the scene coordinate network may adjust its predictions in any way that results in accurate pose estimates.
More formally, we define the following loss function on estimated poses:
\begin{equation}
    \loss^\text{Pose}(\est{\mdl}, \gt{\mdl}) = ||\est{\mathbf{t}}-\gt{\mathbf{t}}|| + \gamma \measuredangle(\est{\boldsymbol{\theta}}, \gt{\boldsymbol{\theta}}),
\end{equation}
with $\mdl = (\boldsymbol{\theta}, \mathbf{t})$ consisting of translation parameters $\mathbf{t}$ and rotation parameters $\boldsymbol{\theta}$.
We denote the ground truth pose parameters as $\gt{\mathbf{t}}$ and $\gt{\boldsymbol{\theta}}$ respectively.
The weighting factor $\gamma$ controls the trade-off between translation and rotation accuracy.
We use $\gamma=100$ in our work, comparing rotation in degree to translation in cm.
Similar to Eq.~\ref{eq:loss:sqrtrgb}, robustify the pose loss by soft clamping.
\begin{equation}
\label{eq:loss:pose}
\est{\loss}^\text{Pose}(\est{\mdl}, \gt{\mdl}) = \begin{cases}
    \loss^\text{Pose}(\est{\mdl}, \gt{\mdl})                                        &\text{if }\loss^\text{Pose}(\est{\mdl}, \gt{\mdl}) < 100\\
    \sqrt{100 \loss^\text{Pose}(\est{\mdl}, \gt{\mdl})}    &\text{otherwise}.
    \end{cases}
\end{equation}
The estimated camera pose $\est{\mdl}$ depends on network parameters $\param$ via the network prediction $\crds$ through robust pose optimization.
In order to optimize the pose loss of Eq.~\ref{eq:loss:pose}, each component involved in pose optimization needs to be differentiable.
In the remainder of this section, we discuss the differentiability of each component and derive approximate gradients where necessary.
We discuss the differentiability of the Kabsch \cite{kabsch1976solution} pose solver for \rgbd images in Sec.~\ref{sec:diffpose:kabsch}.
We give an analytical approximation for gradients of PnP solvers for RGB-based pose estimation in Sec.~\ref{sec:diffpose:pnp}.
In Sec.~\ref{sec:diffpose:refine}, we explain how to approximate gradients of iterative pose refinement.
We discuss differentiable pose scoring via soft inlier counting in Sec.~\ref{sec:diffpose:score}.
Finally, we present a differentiable version of RANSAC, called \emph{differentiable sample consensus} (DSAC) in Sec.~\ref{sec:diffpose:dsac} which also defines our overall training objective.

\subsection{Differentiating Kabsch}
\label{sec:diffpose:kabsch}

We utilize the Kabsch pose solver when estimating poses from \rgbd inputs. 
In this setting, we have 3D-3D correspondences $\set{C}^\text{RGB-D}(\crds)$ given between the 3D coordinates in camera space, defined by the given depth map, and 3D coordinates in scene space $\crds$ predicted by our neural network.
In the following, we assume that we apply the Kabsch solver $g^\text{Kabsch}(\cdot)$ over a subset of correspondences $\set{C}_\set{I}(\crds)$ either when sampling pose hypothesis from three correspondences, or refining the final pose estimate over an inlier set found by RANSAC:
\begin{equation}
    \mdl(\crds) = g^\text{Kabsch}(\set{C}_\set{I}(\crds)) \text{ with } \set{C}_\set{I}(\crds) \subseteq \set{C}^\text{RGB-D}(\crds)
\end{equation}
Here, and in the following, we make the dependence of a model hypothesis to the scene coordinate prediction explicit, \ie we write $\mdl(\crds)$.
The Kabsch solver returns the pose that minimizes the squared residuals over all correspondences:
\begin{equation}
    g^\text{Kabsch}(\set{C}_\set{I}(\crds))  = \argmin_{\mdl'} \sum_{(\eye_i, \crd_i) \in \set{C}_\set{I}(\crds)} r^\text{RGB-D}(\crd_i, \mdl')^2.
\end{equation}
The optimization can be solved in closed form by the following steps \cite{kabsch1976solution}. 
Firstly, we calculate the covariance matrix $\text{cov}[\cdot]$ over the correspondence set:
\begin{equation}
        \text{cov}[\set{C}_\set{I}(\crds)] = \sum_{(\eye_i, \crd_i) \in \set{C}_\set{I}(\crds)} (\eye_i-\bar{\eye})(\crd_i-\bar{\crd})^T,
\end{equation}         
where $\bar{\eye}$ and $\bar{\crd}$ denote the mean over all 3D coordinates in the correspondence set in camera space and scene space, respectively.
Secondly, we apply a singular value decomposition (SVD) to the covariance matrix:
\begin{equation}
        \text{cov}[\set{C}_\set{I}(\crds)] = U \Sigma V^T.
\end{equation}    
We re-assemble the optimal rotation $\boldsymbol{\theta}$, and, subsequently, recover the optimal translation $\mathbf{t}$:
\begin{eqnarray}
    \begin{aligned}        
        &\boldsymbol{\theta} = V \left[ {\begin{array}{ccc}
                   1 & 0 & 0\\
                   0 & 1 & 0\\
                   0 & 0 & \text{det}(VU^T)\\
                  \end{array} } \right] U^T\\
       &\mathbf{t} = \bar{\crd} - \boldsymbol{\theta}(\bar{\eye})\\
       &\mdl(\crds) = (\boldsymbol{\theta}, \mathbf{t}).
    \end{aligned}
\end{eqnarray}
All operations involved in the calculation of $g^\text{Kabsch}(\cdot)$ are differentiable, particularly the gradients of SVD can be calculated according to \cite{Papadopoulo2000diffsvd}, with current deep learning frameworks like PyTorch \cite{paszke2017automatic} offering corresponding implementations.
The differentiability of the Kabsch algorithm has \eg also been utilized in \cite{avetisyan2019ninedof}.

\subsection{Differentiable PnP}
\label{sec:diffpose:pnp}

Similar to the Kabsch solver of the previous section, the PnP solver $g^\text{PnP}(\cdot)$ calculates a pose estimate over a subset $\set{C}_\set{I}(\crds)$ of all correspondences $\set{C}^\text{RGB}(\crds)$, \ie
\begin{equation}
    \mdl(\crds) = g^\text{PnP}(\set{C}_\set{I}(\crds)) \text{ with } \set{C}_\set{I}(\crds) \subseteq \set{C}^\text{RGB}(\crds).
\end{equation}
We utilize a PnP solver when estimating camera poses from RGB images, where 2D-3D correspondences are given between 2D image positions $\pos_i$ and 3D scene coordinate $\crd_i \in \crds$.
A PnP solver optimizes pose parameters to minimize squared re-projection errors:
\begin{equation}
\label{eq:diffpose:pnpobj}
    g^\text{PnP}(\set{C}_\set{I}(\crds))  = \argmin_{\mdl'} ||\mathbf{r}_\set{I}(\crds, \mdl')||^2.
\end{equation}
We construct a residual vector $\mathbf{r}_\set{I}(\cdot)$ over all pixels associated with the current correspondence subset:
\begin{equation}
\label{eq:diffpose:pnpres}
   \mathbf{r}_\set{I}(\crds, \mdl)_i = 
\begin{cases}
    r^\text{RGB}(\crd_i, \mdl)    & \text{if } (\pos_i, \crd_i) \in \set{C}_\set{I}(\crds)\\
    0                                       & \text{otherwise},
\end{cases}
\end{equation}
where $r^\text{RGB}(\cdot)$ denotes a pixels re-projection error, cf.~Eq.~\ref{eq:residual:rgb}.

In contrast to the Kabsch optimization objective, we cannot solve the PnP objective of Eq.~\ref{eq:diffpose:pnpobj} in closed form.
Different PnP solvers have been proposed in the past with different algorithmic structures, \eg \cite{gao2003complete, lepetit2009epnp} or the Levenberg-Marquardt-based optimization in OpenCV \cite{Levenberg44lm, Marquardt63lm,opencv_library}.
Instead of trying to propose a differentiable variant of the aforementioned PnP algorithms, we calculate an analytical approximation of PnP gradients derived from the objective function in Eq.~\ref{eq:diffpose:pnpobj} \cite{Foerstner2016Photogrammetric}.
We have introduced this way of differentiating PnP in the context of neural network training in DSAC++ \cite{brachmann2018lessmore}.

Given a proper initialization, \eg by \cite{gao2003complete, lepetit2009epnp}, we can optimize Eq.~\ref{eq:diffpose:pnpobj} iteratively using the Gauss-Newton method.
Since we are interested only in the gradients of the optimal pose parameters found at the end of optimization, we ignore the influence of initialization itself, avoiding to calculate gradients of complex minimal solvers like \cite{gao2003complete, lepetit2009epnp}.
We give the Gauss-Newton update step to model parameters as 
\begin{equation}
    \mdl^{t+1} = \mdl^t - J_\mathbf{r}^+  \mathbf{r}_\set{I}(\crds, \mdl^t),
\end{equation}
where $J_\mathbf{r}^+ = (J^T_\mathbf{r}J_\mathbf{r})^{-1}J^T_\mathbf{r}$ is the pseudoinverse of the Jacobean matrix $J_\mathbf{r}$ of the residual vector $\mathbf{r}_\set{I}(\crds, \mdl)$ defined in Eq.~\ref{eq:diffpose:pnpres}. 
In particular, the Jacobean matrix $J_\mathbf{r}$ is comprised of the following partial derivatives: 
\begin{equation}
    (J_\mathbf{r})_{ij} = \frac{\partial\mathbf{r}_\set{I}(\crds, \mdl^t)_i}{\partial\mdl^t_j}.
\end{equation}
As mentioned before, the initial pose $\mdl^{t=0}$ may be provided by an arbitrary, non-differentiable PnP algorithm \cite{gao2003complete, lepetit2009epnp}.
We define the pose estimate of the PnP solver as the pose parameters after convergence of the associated optimization problem.
\begin{equation}
    \mdl(\crds) = g^\text{PnP}(\set{C}_\set{I}(\crds)) = \mdl^{t = \infty}
\end{equation}
Thus, we may calculate approximate gradients of model parameters $\mdl(\crds)$ \wrt scene coordinates $\crds$ by fixing the last optimization iteration around the final model parameters:
\begin{equation}
    \derv{\crds}{\mdl(\crds)} \approx - J_\mathbf{r}^+ \derv{\crds}{\mathbf{r}_\set{I}(\crds, \mdl^{t = \infty}}).
\end{equation}

\subsection{Differentiable Refinement}
\label{sec:diffpose:refine}

We refine given camera pose parameters $\mdl$, denoted as $\mathbf{R}(\mdl, \crds)$, by iteratively re-solving for the pose using the set of all inliers $\set{I}$, and updating the set of inliers with the new pose estimate:
\begin{equation}
    \begin{aligned}
    \mdl^{t+1} &= g(\set{C}_{\set{I}^t})\\
    \set{I}^{t+1} &= \{i | r(\crd_i,\mdl^{t+1}) < \tau \}
    \end{aligned}
\end{equation}
We repeat refinement until convergence, \eg when the inlier set $\set{I}$ ceases to change, \ie $\mathbf{R}(\mdl, \crds) = g(\set{C}_{\set{I}^{t=\infty}})$ where $\set{I}^{t=\infty}$ corresponds to the final inlier set.
Similar to differentiating PnP in the previous section, we approximate gradients of iterative refinement by fixing the last refinement iteration.
\begin{equation}
    \derv{\crds}{\mathbf{R}(\mdl, \crds)} = \derv{\crds}{g(\set{C}_{\set{I}^{t=\infty}})},
\end{equation}
where function $g(\cdot)$ denotes either the Kabsch solver or the PnP solver for \rgbd and RGB inputs, respectively. 
We discussed have the calculation of gradients for $g(\cdot)$ already in the previous sections.

\subsection{Differentiable Inlier Count}
\label{sec:diffpose:score}

We obtain a differentiable approximation of inlier counting of Eq.~\ref{eq:hardinliers} by substituting the hard comparison of a pixel's residual to an inlier threshold $\tau$ with a Sigmoid function $\sigma[\cdot]$:
\begin{equation}
    s(\mdl, \crds) = \sum_{\crd_i \in \crds}\sigma[\beta\tau - \beta r(\crd_i, \mdl)].
\end{equation}
For hyper-parameter $\beta$, which controls the softness of the Sigmoid function, we use the following heuristic in dependence of the inlier threshold $\tau$: $\beta = \frac{5}{\tau}$.

\noindent{\textbf{Relation to our Previous Work.}}
In the original DSAC pipeline \cite{brachmann2017dsac} we utilize a designated scoring CNN as a differentiable alternative to traditional inlier counting.
However, our follow-up work on DSAC++ \cite{brachmann2018lessmore} revealed that a scoring CNN is prone to overfitting, and does in general not exceed the accuracy of the simpler soft inlier count.

\subsection{Differentiable RANSAC}
\label{sec:diffpose:dsac}

We can subsume the RANSAC algorithm \cite{ransac1981} in the following three steps, as also discussed in Sec.~\ref{sec:framework}:
Firstly, generate model hypotheses by random sub-sampling of correspondences.
Secondly, choosing the best hypothesis according to a scoring function.
Lastly, refine the winning hypotheses using its inliers.
We discussed the differentiability of most components involved in the previous sub-sections, \eg calculating gradients of pose solvers used for hypothesis sampling and refinement, and differentiating the inlier count for hypothesis scoring.
However, choosing the best hypothesis according to Eq.~\ref{eq:framework:ransac} involves a non-differentiable $\argmax$ operation. 

In \cite{brachmann2017dsac}, we introduce a differentiable approximation of hypothesis selection in RANSAC, called \emph{differentiable sample consensus} (DSAC).
In \cite{brachmann2017dsac}, we also argue, and show empirically, that a simple soft $\argmax$ approximation, a weighted average of arguments, does not work well.
A soft $\argmax$ can be unstable when arguments have multi-modal structure, \ie very different arguments have high weights in the average.
A standard average might also be overly sensitive to outlier arguments.

The DSAC approximation relies on a probabilistic selection of a model hypothesis according to a probability distribution $ p(j|\crds)$ over the discrete set of sampled hypotheses with index $j$:
\begin{equation}
\begin{aligned}
    \tilde{\mdl}(\crds) &= \mdl_j(\crds)~\text{with}~j\sim p(j|\crds)\\
    \est{\mdl}(\crds) &= \refine(\tilde{\mdl}(\crds), \crds),
    \end{aligned}
\end{equation}
where $\tilde{\mdl}(\crds)$ denotes the selected hypothesis, and $\est{\mdl}(\crds)$ denotes the final, refined estimate of our pipeline.
The distribution guiding hypothesis selection is a softmax distribution over scores, \ie
\begin{equation}
    \label{eq:diffpose:distribution}
    p(j|\crds) = \frac{\exp [\alpha  s(\mdl_j(\crds), \crds)]}{\sum_{k=1}^M \exp [\alpha s(\mdl_k(\crds), \crds)] }.
\end{equation}
The hyper-parameter $\alpha$ corresponds to a temperature that controls the softness of the distribution. 
The larger $\alpha$, the more DSAC will behave like RANSAC in always selecting the hypothesis with maximum score, while providing less signal for learning.
In DSAC++ \cite{brachmann2018lessmore}, we present a schema do adjust $\alpha$ automatically during learning.
In this work, we treat $\alpha$ as a hand-tuned and fixed hyper-parameter, as we found the camera re-localization problem not overly sensitive to the exact value of $\alpha$, and fixing it simplifies the software architecture of our pipeline.
 
\begin{figure}[!t]
    \centering
    \includegraphics[width=1\linewidth]{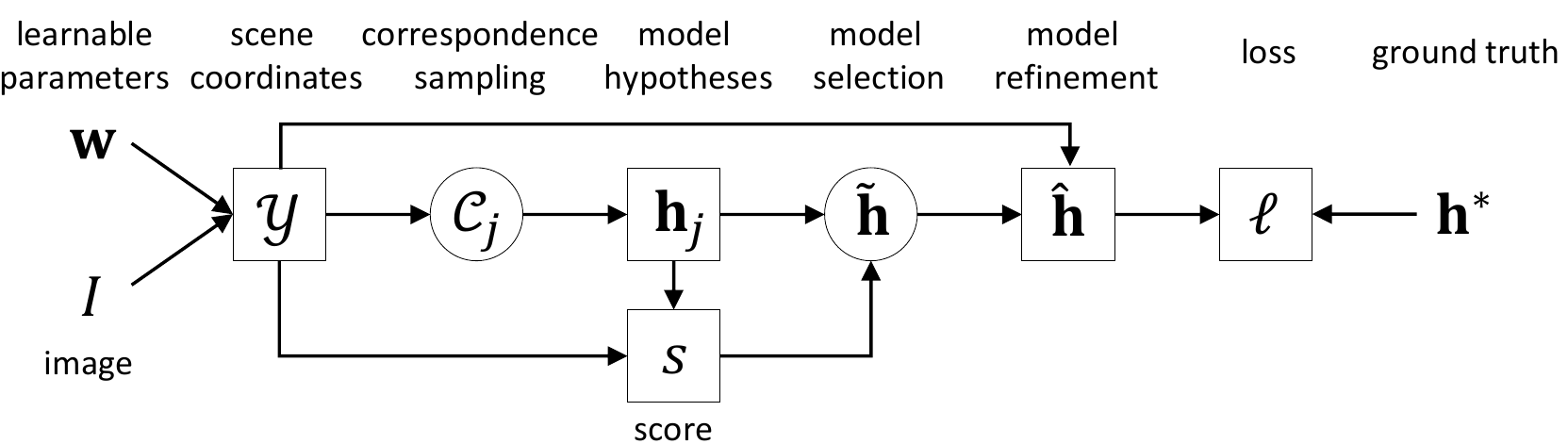}
    \caption{\textbf{DSAC Computation Graph \cite{compgraph2015}.} Nodes without frames represent inputs to the system, nodes with square frames represent deterministic operations, nodes with circular frames represent sampling operations. Arrows denote an input relation.}
    \label{fig:method:graph}
\end{figure} 
 
To learn the pipeline, we optimize the expectation of the pose loss $\est{\loss}^\text{Pose}$ of Eq.~\ref{eq:loss:pose} \wrt randomly selecting hypotheses:
\begin{equation}
    \label{eq:loss:exp}
    \Loss^\text{Pose}(\crds, \gt{\mdl}) = \expectation{j\sim p(j|\crds)}{\est{\loss}^\text{Pose}(\refine(\cdot), \gt{\mdl})},
\end{equation}
where we abbreviate the final, refined camera pose $\refine(\mdl_j(\crds), \crds)$ as $\refine(\cdot)$.
To minimize the expectation, the neural network should learn to predict scene coordinates $\crds$ that ensure the following two properties:
Firstly, hypotheses with a large loss after refinement should receive a low selection probability, \ie a low soft inlier count.
Secondly, hypotheses with a high soft inlier count should receive a small loss after refinement.
We present a schematic overview of all components involved in our DSAC-based pipeline in Fig.~\ref{fig:method:graph}.
The figure summarises dependencies between processing steps, and differentiates between deterministic functions and sampling operations.
The graph structure illustrates the non-trivial relation between the scene coordinate prediction and pose quality, since scene coordinates directly influence pose hypotheses, scoring and refinement.

The DSAC training objective of Eq.~\ref{eq:loss:exp} is smooth and differentiable, and its gradients can be formulated as follows:
\begin{equation}
    \label{eq:loss:grad}
    \derv{\crds}{\Loss^\text{Pose}(\cdot)} = \expectation{j}{\est{\loss}^\text{Pose}(\cdot)\derv{\crds}{\log p(j|\crds)} + \derv{\crds}{\est{\loss}^\text{Pose}(\cdot)}},
\end{equation}
where we use $\cdot$ as a stand-in for the respective function arguments in Eq.~\ref{eq:loss:exp}, and abbreviate the expectation over $p(j|\crds)$ as $\expectation{j}{\cdot}$.
We use Eq.~\ref{eq:loss:grad} to learn our system in an end-to-end fashion, updating neural network parameters $\param$ of scene coordinate prediction $\crds = f(I,\param)$.

\section{Experiments}
\label{sec:experiments}

We evaluate our camera re-localization pipeline for two indoor datasets and one outdoor dataset.
Firstly, in Sec.~\ref{sec:exp:setup} we discuss our experimental setup, including datasets, training schedule, hyper-parameters and competitors.
Secondly, we report results on 3 different datasets in Sections \ref{sec:exp:7scenes}, \ref{sec:exp:12scenes} and \ref{sec:exp:cambridge}, respectively.
Thirdly, we provide several ablation studies in Sections \ref{sec:exp:network}, \ref{sec:exp:aug}, \ref{sec:exp:rf} and \ref{sec:exp:end2end}, as well as visualizations of scene representations learned by our system in Sec.~\ref{sec:exp:geometry}.
Furthermore, we analyze the scene compression properties of DSAC* in Sec.~\ref{sec:exp:compression}.

\subsection{Setup}
\label{sec:exp:setup}

\noindent{\textbf{Task Variants.}}
We deploy our system in several flavours, catering to different application scenarios where depth measurements or 3D scans of a scene might be available or not.
Specifically, we analyze the following settings:
\begin{itemize}
    \item \textbf{RGB-D:} We have \rgbd images for training as well as at test time. 
    For initialization training, we render ground truth scene coordinates using 3D scans of each scene. 
    For end-to-end training and at test time, we generate camera coordinates $\eye$ from the \rgbd depth channels.
    We use a Kabsch \cite{kabsch1976solution} pose solver for sampling hypotheses and for refining the final estimate.
    \item \textbf{RGB + 3D model:} We have RGB images for training as well as at test time. We can render ground truth scene coordinates for training using a 3D model of the scene. The 3D model can either be a sparse SfM point cloud, or a dense 3D scan. We use the PnP solver of Gao \etal \cite{gao2003complete} to sample camera pose hypotheses, and the Levenberg-Marquardt \cite{Levenberg44lm, Marquardt63lm} PnP optimizer of OpenCV \cite{opencv_library} for final refinement.
    \item \textbf{RGB:} Same as the previous setting, but we have no information about the 3D geometry of a scene, only RGB images and ground truth poses for training. To initialize scene coordinate regression, we optimize the heuristic objective of Eq.~\ref{eq:loss:rgbd}.
\end{itemize}

\noindent{\textbf{Hyper-Parameters.}}
We convert input images to grayscale and re-scale them to 480px height.
For training, we follow Li \etal \cite{li2020hierarchical} and apply data augmentation.
We apply random adjustments of brightness and contrast of the input image within a $\pm10\%$ range.
We randomly rotate images, ground truth scene coordinates and camera poses within a $\pm30^\circ$ range.
We randomly re-scale images within 66$\%$ and 150$\%$, and adjust the camera focal length accordingly.
Different from Li \etal \cite{li2020hierarchical}, we do not shear training images, since our simple pinhole camera models does not support this operation. 
We also do not shift training images, since for a patch-based network architecture, shifting by more than 4px would just increase the period of the input without any effect other than increasing boundary effects.

We use an inlier threshold $\tau=10\text{px}$ for RGB-based pose optimization, and $\tau=10\text{cm}$ for RGB-D-based pose optimization.
We sample $M=64$ RANSAC hypotheses. 
We reject an hypothesis if the corresponding minimal set of scene coordinates does not satisfy the inlier threshold \cite{chum2002tdd}, and sample again.
We score hypotheses using a soft inlier count at training and test time. 
For training, we optimize the expectation over hypothesis selection according to the distribution of Eq.~\ref{eq:diffpose:distribution} with a temperature of $\alpha=\frac{100}{|\crds|}$, where $|\crds|$ corresponds to the number of scene coordinates predicted, resp.~to the output resolution of the neural network.
At test time, we resort to standard RANSAC, and choose the best hypothesis with highest score.
We do at most 100 refinement iterations, but stop early if the inlier set converges which typically takes at most 10 iterations.

We initialize the scene coordinate network for 1M iterations, a batch size of 1 image, and the Adam optimizer \cite{adam2014} with a learning rate of $10^{-4}$. 
This stage takes approximately two days on a single Tesla K80 GPU.
We train the system end-to-end for another 100k iterations, and a learning rate of $10^{-6}$, which takes 12 hours on the same hardware. 
Our implementation, based on PyTorch \cite{paszke2017automatic}, is publicly available: \url{https://github.com/vislearn/dsacstar}.

\noindent{\textbf{Datasets.}}
We evaluate our pipeline on three standard camera re-localization datasets, both indoor and outdoor:
\begin{itemize}
    \item \textbf{7Scenes\cite{shotton13scorf}:} A \rgbd indoor re-localization dataset of seven small indoor environments featuring difficult conditions such as motion blur, reflective surfaces, repeating structures and texture-less areas.
    Images were recorded using KinectFusion \cite{izadi2011kinectfusion} which also provides ground truth camera poses. 
    For each scene, several thousand frames are available which the authors split into training and test sets.
    The depth channels of this dataset are not registered to the color images. 
    We register them by projecting the depth maps to 3D points using the depth sensor calibration, and re-projecting them using the color sensor calibration while taking the relative transformation between depth and color sensor into account.
    A dense 3D scan of each scene is available for rendering ground truth coordinates for initialization training.
    \item \textbf{12Scenes\cite{valentin2016learning}:} A \rgbd indoor re-localization dataset similar to 7Scenes, but containing twelve slightly larger indoor environments. 
    Each scene comes with several hundred frames, split by the authors into training and test sets.
    The depth maps provided by the authors are registered to the color images.
    A dense 3D scan of each scene is available as well, which we use to render ground truth scene coordinates for initialization training.
    \item \textbf{Cambridge\cite{kendall2015convolutional}}: A RGB outdoor re-localization dataset of five landmarks in Cambridge, UK. Compared to the previous indoor datasets, each landmark spans an area of several hundred or thousand square meters.
    Each scene comes with several hundred frames, split by the authors into training and test tests. The authors also provide ground truth camera poses re-constructed using a SfM tool. The sparse SfM point cloud is also available for each scene, which we use to render sparse scene coordinate ground truth for RGB-based re-localization. The dataset contains a sixth scene, an entire street scene, which me omit in our experiments. The corresponding reconstruction is of poor quality containing several outlier camera poses and 3D points as well as duplicated and diverging geometry. Like in our previous work \cite{brachmann2018lessmore,brachmann2019neural} we were unable to achieve reasonable accuracy on the street scene.
\end{itemize}

\noindent{\textbf{Competitors.}}
We compare to the following \textbf{absolute pose regression} networks:
PoseNet (the updated version of 2017) \cite{geometricloss}, SpatialLSTM \cite{LSTMPoseNet}, MapNet \cite{mapnet2018} and SVS-Pose \cite{naseer2017svspose}.
We compare to the following \textbf{relative pose estimation} approaches:
AnchorNet \cite{saha2018anchornet}, and retrieval-based InLoc \cite{taira2018inloc}.
For feature-based competitors, we report results of the ORB baseline used in \cite{shotton13scorf} and \cite{valentin2016learning}, as well as the SIFT baseline used in \cite{valentin2016learning}.
For a state-of-the-art feature-based pipeline, we compare to ActiveSearch \cite{sattler2016efficient}.
Several early scene coordinate regression works were based on \textbf{random forests}. 
We compare to SCoRF of Shotton \etal \cite{shotton13scorf}, and its extension to multi-output forests (\emph{MO Forests}) \cite{guzman2014multi} and forests predicting Gaussian mixture models (GMM) of scene coordinates, in the variation of Valentain \etal \cite{valentin2015cvpr} for \rgbd (\emph{GMM F. (V)}) and of Brachmann \etal for RGB (\emph{GMM F. (B)}).
Furthermore, we compare to the Back-Tracking Forests of Meng \etal \cite{meng17} (\emph{BTBRF}), to the Point-Line Forests of Meng \etal \cite{meng18} (\emph{PLForests}), and MNG forests \cite{valentin2016learning}.
For \textbf{CNN-based} scene coordinate regression, we compare to ForestNet\cite{rfvscnn2016}, scene coordinate regression with an angle-based loss \cite{li2018abrloss} (\emph{ABRLoss}), joint scene coordinate classification and regression \cite{li2020hierarchical} (\emph{SCoCR}) and the visual descriptor learning approach of Schmidt \etal \cite{schmidt2017ssvdl} (\emph{SS-VDL}).
We also include results of the adaptive forests of Cavallari \etal \cite{cavallari2017fly} (\emph{OtF Forests}), comparing to their \emph{headline performance}. 
Finally, we compare to previous iterations of this pipeline, namely DSAC \cite{brachmann2017dsac} and DSAC++ \cite{brachmann2018lessmore}. 
We denote our updated pipeline, described in this article, as DSAC*.

\subsection{Results for Indoor Localization (7Scenes)}
\label{sec:exp:7scenes}

\begin{figure*}[!t]
    \centering
    \includegraphics[width=0.9\linewidth]{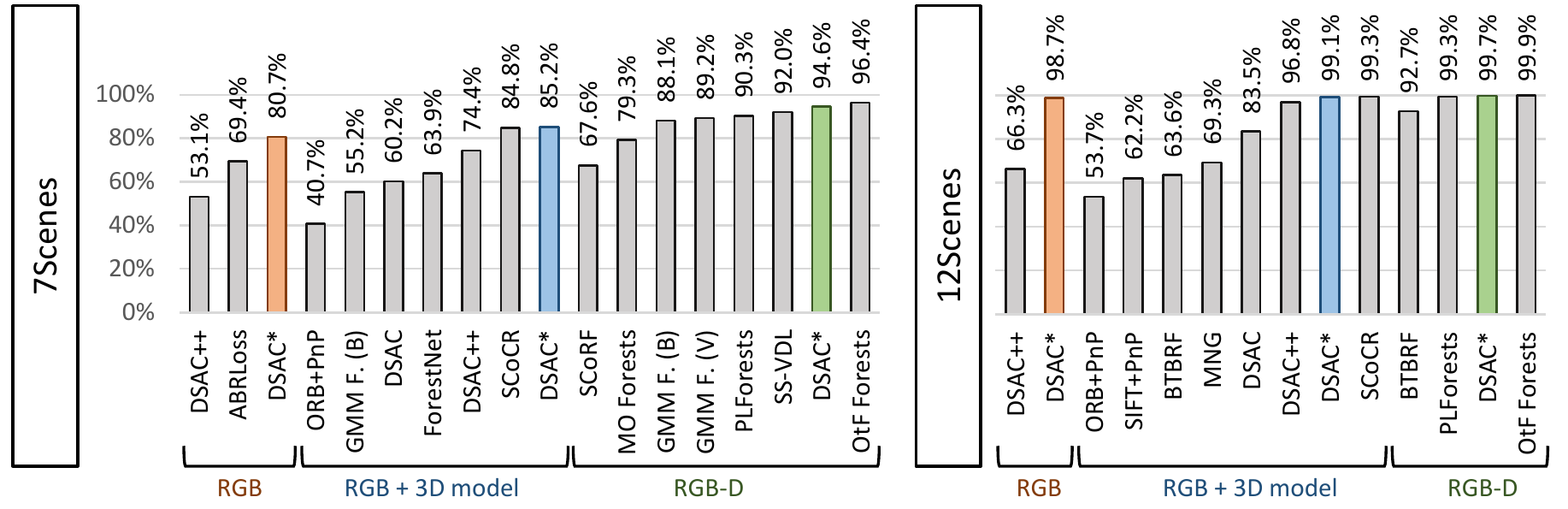}
    \vspace{-0.5cm}
    \caption{\textbf{Indoor Localization Accuracy.} We report the average percentage of correctly re-localized frames below an error threshold of 5cm and 5$^\circ$ on the 7Scenes \cite{shotton13scorf} and 12Scenes \cite{valentin2016learning} datasets. We group methods by utilized data, \ie \textbf{RGB}: neither a 3D model or depth maps at training and test time, \textbf{RGB + 3D model}: a 3D model or depth maps at training time \emph{but not at test time}, \textbf{RGB-D}: depth maps at training time \emph{and at test time}. See the main text for references to all methods.}
    \label{fig:exp:scenes_bars}
    \vspace{-0.3cm}
\end{figure*}

We train one scene coordinate regression network per scene, and accept a pose estimate for a test image if its pose error is below 5$^\circ$ and 5cm.
We calculate the accuracy per scene, and report the average accuracy over all 7Scenes, see quantitative results in Fig.~\ref{fig:exp:scenes_bars}, left.

\noindent{\textbf{RGB.}}
For training from RGB images and ground truth poses only, our new training procedure and network architecture increases accuracy significantly compared to DSAC++ (+27.6\%). 
DSAC* also achieves higher accuracy than the angle-based loss of Li \etal \cite{li2018abrloss}, despite the latter incorporating multi-view constraints and a photometric loss.
We attribute some (but not all) of the performance gain to using training data augmentation, see Sec.~\ref{sec:exp:aug} for a discussion.

\noindent{\textbf{RGB + 3D model.}} 
When a 3D model is available to render ground truth scene coordinates for training, both DSAC++ and DSAC* benefit, with DSAC* achieving highest accuracy with 85.2\% of re-localized frames. 
Also note that DSAC* is trained for 2.5 days compared to 6 days for DSAC++ on identical hardware.
\emph{SCoCR} \cite{li2020hierarchical} achieves similar accuracy but leverages a more complicated network architecture with multiple, hierarchical classification heads conditioning the scene coordinate regression head as well as higher model capacity (165MB vs.~28MB). 
Note that \emph{SCoCR} deploys training data augmentation similar to our setup.

\noindent{\textbf{RGB-D.}}
When DSAC* estimates poses from \rgbd images, it achieves 
accuracy comparable to the state-of-the-art approach \emph{OtF Forests} of Cavallari \etal \cite{Cavallari2019cascade}.  
Cavallari \etal \cite{Cavallari2019cascade} use an ICP and rendering-based post-processing to achieve their top-accuracy. 
Without post-processing, they report 93.4\% on 7Scenes, slightly lower than the accuracy of DSAC*.
Note that the difference in accuracy for DSAC* compared to the RGB setups solely stems from the use of Kabsch as a pose solver, since our network still estimates scene coordinates from a grayscale image.
The correct depth of image points allows a re-localization pipeline to trivially infer the distance between camera and scene.
Note that all RGB-D competitors model the uncertainty of scene coordinates in some form, \ie predicting full distributions of image-to-scene correspondences. 
Compared to this, the expressiveness of our framework is limited by only predicting scene coordinate point estimates. 

\begin{figure*}[!t]
    \centering
    \includegraphics[width=1\linewidth]{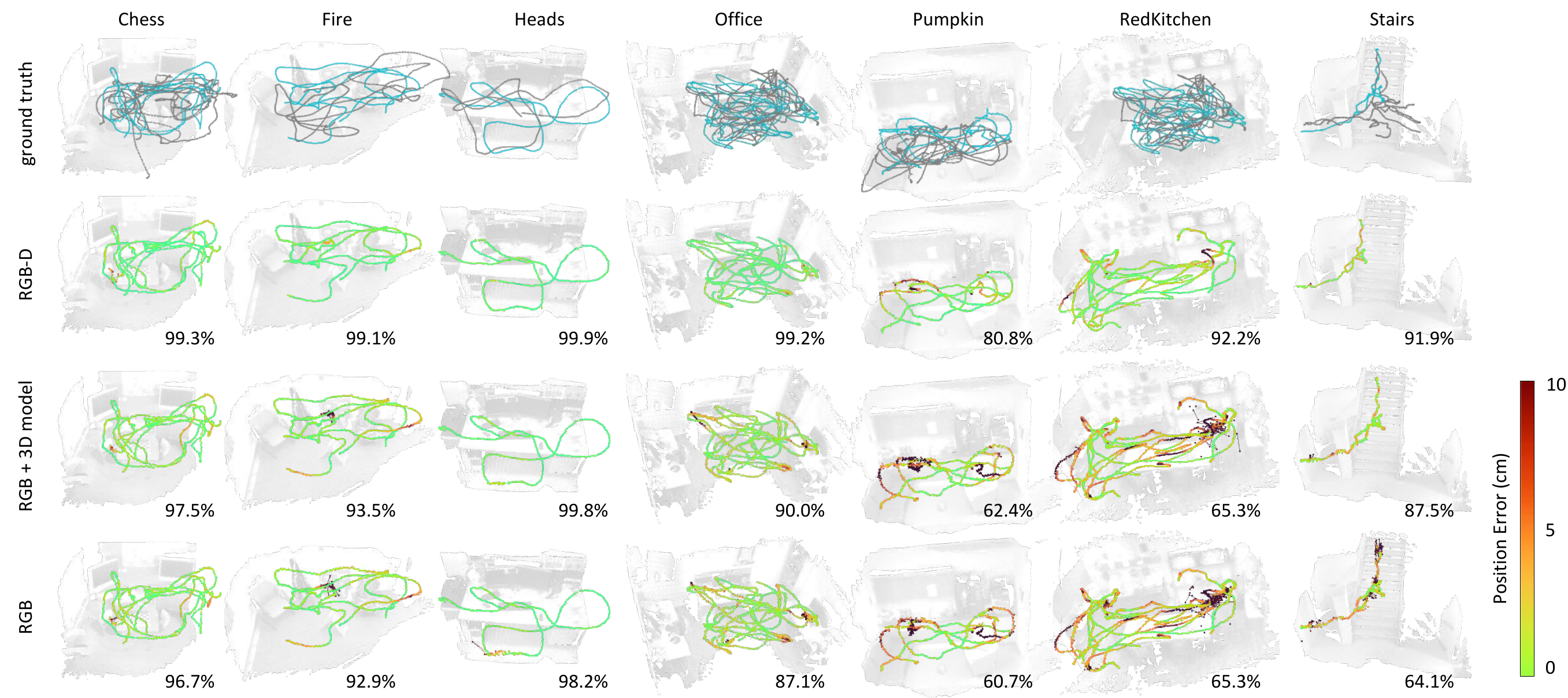}
    \vspace{-0.3cm}
    \caption{\textbf{Results For Indoor Scenes. First Row:} Camera positions of training frames in {\color{gray}gray} and of test frames in {\color{cyan}cyan} for all scenes of the 7Scenes \cite{shotton13scorf} dataset. \textbf{Remaining Rows:} Estimated camera positions of test frames, color coded by position error. We also state the percentage of test frames with a pose error below 5cm and 5$^\circ$. Each row represents a different training setup. For a more informative visualization, we show the ground truth 3D scene model as a faint backdrop, and we connect consecutive frames within 50cm tolerance.}
    \vspace{-0.3cm}
    \label{fig:exp:7scenes_paths}
\end{figure*}

\begin{figure*}[!t]
    \centering
    \includegraphics[width=1\linewidth]{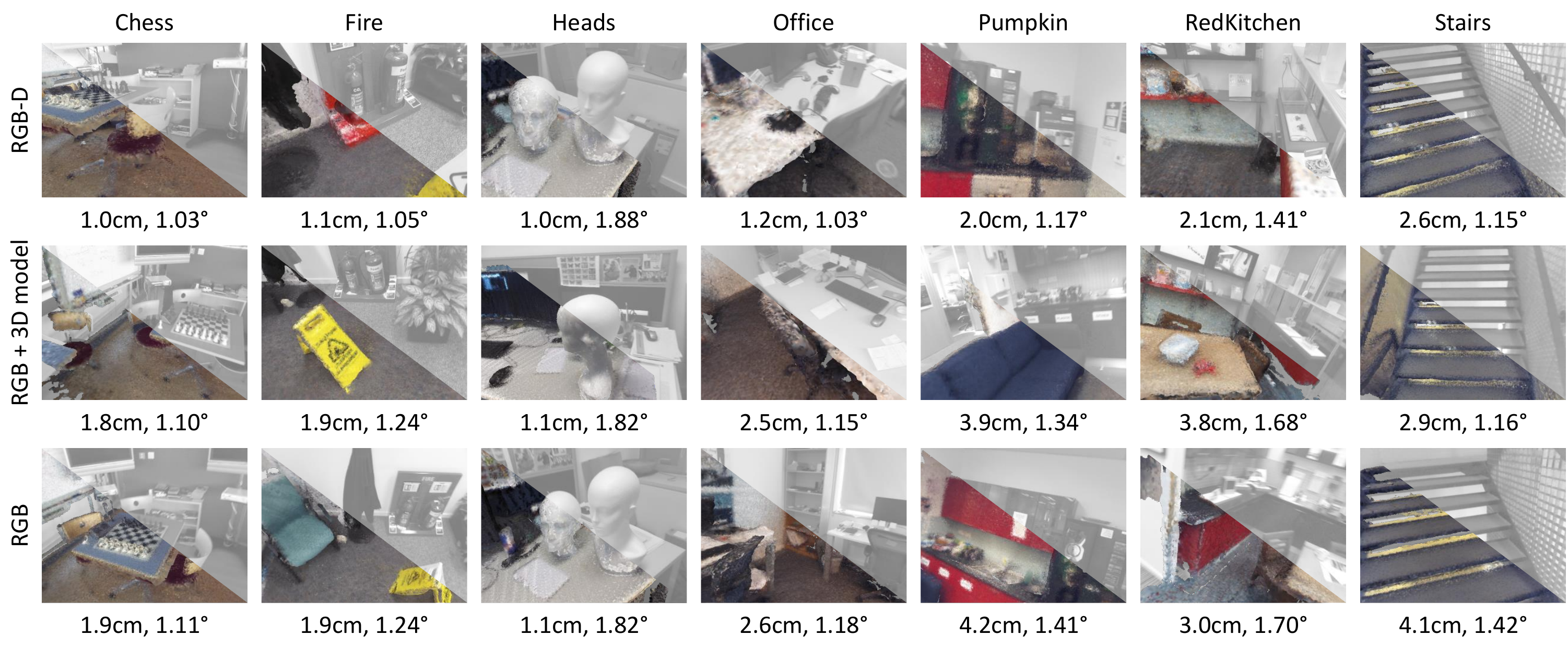}
    \caption{\textbf{Median Errors for Indoor Scenes.} For all test sequences of the 7Scenes \cite{shotton13scorf} dataset, we select the frame with the median pose estimation error. We show the original input frame in gray scale, and a rendered overlay in color using the estimated pose, and the ground truth 3D model. We write the associated median pose error below each instance.  Each row represents a different training setup.}
    \label{fig:exp:7scenes_median}
\end{figure*}

\noindent{\textbf{Qualitative Results.}}
We visualize the estimated test trajectory, as well as the pose error, of DSAC* for all scenes and all re-localization settings in Fig.~\ref{fig:exp:7scenes_paths}.
Estimated trajectories are predominately smooth, with outlier predictions concentrated on particular, presumably difficult, areas of each scene.
To also visualize the re-localization quality in an augmented reality setup, we compare renderings of 3D models of each scene, using estimated camera poses, with the associated test image in Fig.~\ref{fig:exp:7scenes_median}.
To give an unbiased impression of the general re-localization quality, we selected the test frame with median pose error for each visualization.

\subsection{Results for Indoor Localization (12Scenes)}
\label{sec:exp:12scenes} 

We report quantitative results for 12Scenes in Fig.~\ref{fig:exp:scenes_bars}, right.
DSAC* achieves state-of-the-art accuracy in all settings for this dataset, consistently outperforming DSAC++. 
In general, we would consider this dataset being solved, with multiple methods achieving an average accuracy of $\approx$99\% for re-localization from RGB-D and RGB images, and in case of DSAC* even from RGB images without a 3D model, the most difficult setting.

\subsection{Results for Outdoor Localization (Cambridge)}
\label{sec:exp:cambridge} 

\begin{table*}[!t]
\renewcommand{\arraystretch}{1.3}
\caption{\textbf{Outdoor Localization Accuracy.} We report median errors on the Cambridge Landmarks\cite{kendall2015convolutional} dataset as translation error (cm) / rotation error ($^\circ$). \emph{\color{gray}{N/A}} denotes that a particular result was not reported, whereas a dash (-) indicates that a method does not support this particular setting, \ie training with or without a 3D model, respectively. Best results in \textbf{bold} per column, second best \underline{underlined}. For DSAC variants we additionally state, in brackets, the training time in days on identical hardware.}
\label{tab:exp:cambridge}
\centering
\begin{tabular}{lc||ccccc||ccccc}
              & & \multicolumn{5}{c||}{RGB + 3D model}               & \multicolumn{5}{c}{RGB}                          \\
Method        & & Church  & Court   & Hospital & College & Shop    & Church  & Court   & Hospital & College & Shop    \\ \hline
MapNet \cite{mapnet2018}      & &    -     &    -     &     -     &     -    &      -   & 200/4.5 & \color{gray}{N/A}     & 194/3.9  & 107/1.9 & 149/4.2 \\
SpatialLSTM \cite{LSTMPoseNet}  &  &    -     &     -    &     -     &   -      &     -    & 152/6.7 & \color{gray}{N/A}     & 151/4.3  & 99/1.0  & 118/7.4 \\
SVS-Pose \cite{naseer2017svspose}    &  &    -     &     -    &      -    &     -    &     -    & 211/8.1 & \color{gray}{N/A}     & 150/4.0  & 106/2.8 & 63/5.7  \\
PoseNet17 \cite{geometricloss}      & & 149/3.4 & 700/3.7 & 217/2.9  & 99/1.1  & 105/4.0 & 157/3.2 & 683/3.5 & 320/3.3  & 88/1.0  & 88/3.8  \\
AnchorNet \cite{saha2018anchornet}   &  &     -    &     -    &      -    &    -     &     -   & 104/2.7 & \color{gray}{N/A}     & 121/2.6  & 57/0.9  & 52/2.3  \\
InLoc \cite{taira2018inloc}     &    & 18/0.6  & 120/0.6 & 48/1.0   & 46/0.8  & 11/0.5  &    -     &    -     &    -      &      -   &      -   \\
Active Search \cite{sattler2016efficient} & & 19/0.5  & \color{gray}{N/A}     & 44/1.0   & 42/0.6  & 12/0.4  &    -     &     -    &    -      &     -    &      -   \\
BTBRF \cite{meng17}     &    & 20/0.4  & \color{gray}{N/A}     & 30/0.4   & 39/0.4  & 15/0.3  &    -     &     -    &     -     &   -      & -        \\
SANet \cite{yang2019sanet}       &  & 16/0.6  & 328/2.0 & 32/0.5   & 32/0.5  & 10/0.5  &    -     &     -    &     -     &    -     &      -   \\  \hline
DSAC \cite{brachmann2017dsac}   & (4d) & 55/1.6  & 280/1.5 & 33/0.6   & 30/0.5  & 9/0.4   &    -     &   -      &     -     &     -    &    -     \\
DSAC++ \cite{brachmann2018lessmore}     & (6d) & \underline{13/0.4}  & \underline{40/0.2}  & \textbf{20/0.3}   & 18/0.3  & \underline{6/0.3}   & \underline{20/0.7}  & \underline{66/0.4}  & \underline{24/0.5}   & \underline{23/0.4}  & \underline{9/0.4}   \\
NG-DSAC++ \cite{brachmann2019neural} & (6d) & \textbf{10/0.3}  & \textbf{35/0.2}  & 22/0.4   & \textbf{13/0.2}  & \underline{6/0.3}   & \color{gray}{N/A}     & \color{gray}{N/A}     & \color{gray}{N/A}      & \color{gray}{N/A}     & \color{gray}{N/A}     \\
DSAC*   & (2.5d) & \underline{13/0.4}  & 49/0.3  & \underline{21/0.4}   & \underline{15/0.3}  & \textbf{5/0.3}   & \textbf{15/0.6}  & \textbf{34/0.2}  & \textbf{21/0.4}   & \textbf{18/0.3}  & \textbf{5/0.3}      
\end{tabular}
\end{table*}

We measure the re-localization quality on the Cambridge dataset using the median pose error for each scene, 
see Table \ref{tab:exp:cambridge}.
Due to the ground truth for this dataset being recovered using a SfM tool, we report results with centimeter precision. 
We find the expressiveness of millimeter precision dubious given the nature of ground truth poses.
Given a 3D model for training, DSAC* and DSAC++ achieve similar accuracy, but DSAC* trains significantly faster.
For many scenes, NG-DSAC++ \cite{brachmann2019neural}, \ie DSAC++ with neural-guided RANSAC, achieves best results.
In principle, we could extend DSAC* to utilize neural guidance as well.
Neural guidance is designed to improve RANSAC in high outlier domains.
We expect the benefit of coupling it with DSAC* to be rather small, given the quality of results already.

When training without a 3D model, the new training objective of DSAC* achieves higher accuracy than DSAC++ across all scenes.
Notably, DSAC* trained without a 3D model achieves higher accuracy than any method (including DSAC*) trained with a 3D model for the Great Court scene. 
Great Court is the largest landmark in the dataset. 
The associated SfM reconstruction contains a high outlier ratio, and might hinder the training process due to its low quality.

\begin{figure*}[!t]
    \centering
    \includegraphics[width=1\linewidth]{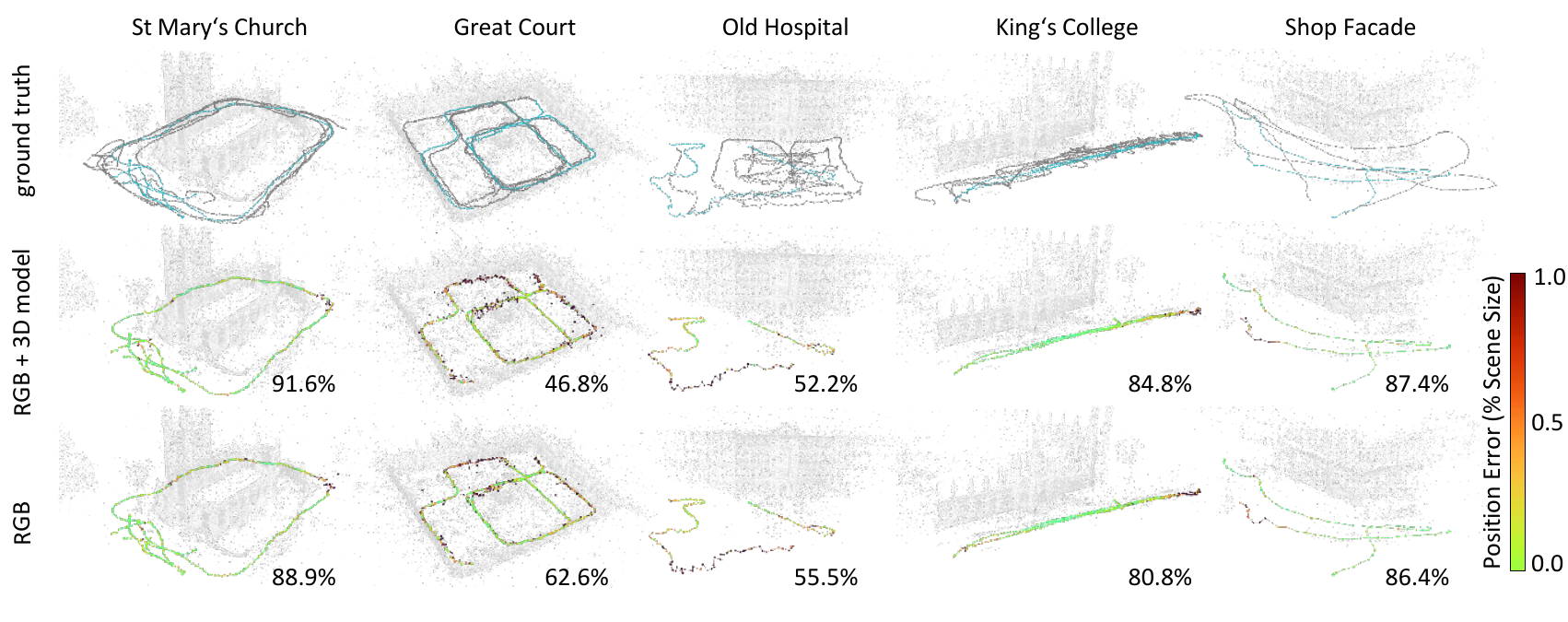}
    \caption{\textbf{Results For Outdoor Scenes. First Row:} Camera positions of training frames in {\color{gray}gray} and of test frames in {\color{cyan}cyan} for scenes of the Cambridge Landmarks \cite{kendall2015convolutional} dataset. \textbf{Remaining Rows:} Estimated camera positions of test frames, color coded by position error. We also state the percentage of test frames with a position error below 0.5\% of the scene size. We derive the threshold for each scene from the scene extent given in \cite{kendall2015convolutional}. In particular, we use 35cm for St. Mary's Church, 45cm for Great Court, 22cm for Old Hospital, 38cm for King's College and 15cm for Shop Facade. Each row represents a different training setup. For a more informative visualization, we show the ground truth 3D scene model as a faint backdrop, and we connect consecutive frames within 5m tolerance.}
    \label{fig:exp:cambridge_paths}
\end{figure*}

\begin{figure*}[!t]
    \centering
    \includegraphics[width=1\linewidth]{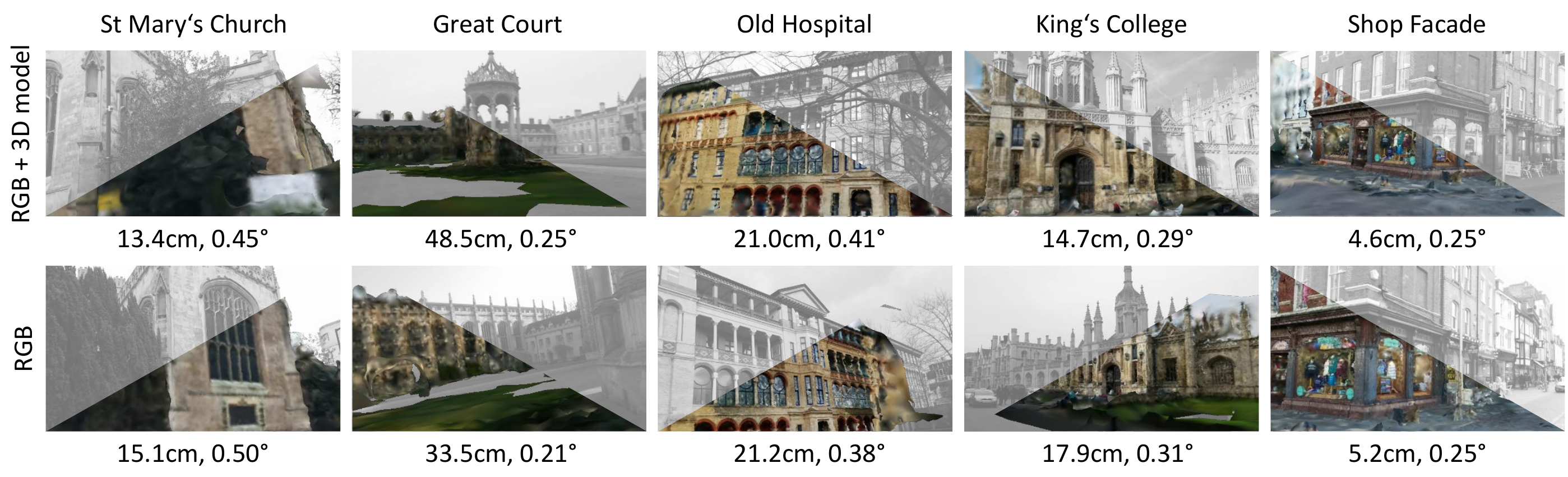}
    \vspace{-0.5cm}
    \caption{\textbf{Median Errors for Outdoor Scenes.} For all test sequences of the Cambridge Landmarks \cite{kendall2015convolutional} dataset, we select the frame with the median pose estimation error. We show the original input frame in gray scale, and a rendered overlay in color using the estimated pose, and a 3D scene model generated from the ground truth structure-from-motion point cloud. We write the associated median pose error below each instance.  Each row represents a different training setup.}
    \label{fig:exp:cambridge_median}
    \vspace{-0.5cm}
\end{figure*}

We visualize the estimated test trajectories of DSAC* in Fig.~\ref{fig:exp:cambridge_paths}. 
Due to the very different scene sizes, we derive a scene-dependent threshold to color-code pose errors.
The visualizations reveal that high localization error is correlated with the distance of the camera to the scene, particularly obvious for Old Hospital, but also King's College.
In Fig.~\ref{fig:exp:cambridge_median}, we depict the median pose error per scene, and observe a high visual quality of re-localization, suitable for augmented reality applications.

\subsection{Network Architecture and Runtime} 
\label{sec:exp:network}

\begin{table}[]
\centering
\caption{\textbf{Comparison of Network Architecture.} We compare statistics of the network architecture of DSAC* (this work) and DSAC++ \cite{brachmann2018lessmore}. We train both architectures using the new training schedule of DSAC* (2.5 days). We report accuracy on the 7Scenes dataset for the \emph{RGB + 3D model} setting.}
\label{tab:architecture}
\begin{tabular}{@{}lccclc@{}}
\toprule
Architecture & Size & Time & RF & \begin{tabular}[l]{@{}l@{}}Training \\ Procedure\end{tabular} & Accuracy \\ \midrule
\multirow{2}{*}{\begin{tabular}[c]{@{}l@{}}DSAC++\\ (VGGNet)\end{tabular}} & \multirow{2}{*}{104MB} & \multirow{2}{*}{150ms} & \multirow{2}{*}{73px} & DSAC++ & 74.4\% \\ \cmidrule(l){5-6} 
 &  &  &  & DSAC* & 82.0\% \\ \midrule
\multirow{2}{*}{\begin{tabular}[c]{@{}l@{}}DSAC*\\ (ResNet)\end{tabular}} & \multirow{2}{*}{28MB} & \multirow{2}{*}{50ms} & \multirow{2}{*}{81px} & \multirow{2}{*}{DSAC*} & \multirow{2}{*}{85.2\%} \\
 &  &  &  & &  \\ \bottomrule
\end{tabular}
\end{table}

As explained in Sec.~\ref{sec:score}, we updated the network architecture compared to DSAC++. 
To disambiguate the impact of the network, and of the updated training schedule, we conduct an ablation study, see Table.~\ref{tab:architecture}.
We trained both architectures using the updated training schedule of DSAC* on the 7Scenes dataset.
Both architectures benefit from the DSAC* training settings.
However, the DSAC* architecture combined with DSAC* training achieves best accuracy, despite faster run time and smaller memory footprint.
Together with the stream-lined pose optimization (\eg using 64 RANSAC hypotheses instead of 256), we achieve a total runtime of the system of 75ms compared to 200ms for DSAC++ on a single Tesla K80 GPU.

\subsection{Impact of Data Augmentation} 
\label{sec:exp:aug}

\begin{figure*}[!t]
    \centering
    \includegraphics[width=1\linewidth]{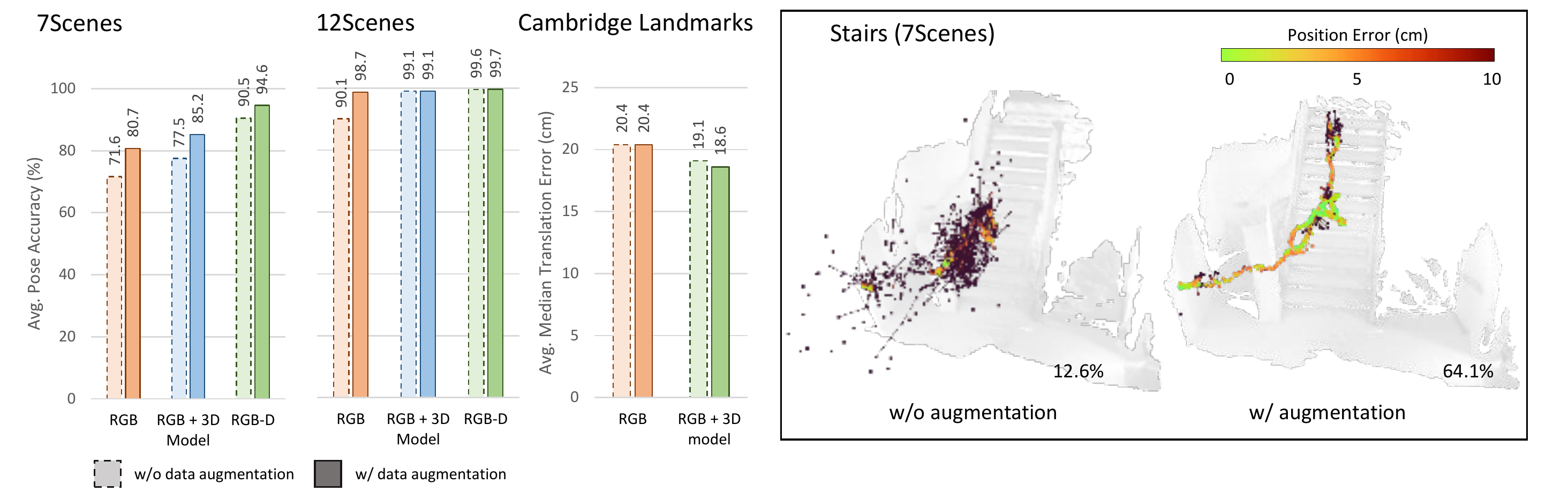}
    \caption{\textbf{Effect of Data Augmentation. Left:} Average percentage of correctly localized frames on 7Scenes\cite{shotton13scorf} and 12Scenes\cite{valentin2016learning}, as well as average median translations errors for Cambridge Landmarks\cite{kendall2015convolutional}, with and without using geometric training data augmentation (random rotation and scaling). \textbf{Right:} We show qualitative results on the 7Scenes Stairs sequence when training in RGB-only mode (\ie without using a 3D model of the scene).}
    \label{fig:exp:aug}
\end{figure*}

Li \etal \cite{li2020hierarchical} demonstrated the effectiveness of simple geometric training data augmentation (random rotation and re-scaling) for RGB-based camera re-localization (w/ 3D model) on the 7Scenes dataset. 
We confirm these results here, and show similar effects for RGB-D based re-localization, and RGB-only re-localization. 
See Fig.~\ref{fig:exp:aug} for quantitative and qualitative results. 
Data augmentation results in significant improvement for DSAC* on the 7Scenes dataset with +9.1\%, +7.7\% and 4.1\% improvement depending on the setup.
We see largest improvements on the 7Scenes Stairs sequence when training in RGB-only mode (+51.5\%), see Fig.~\ref{fig:exp:aug}, right.
This scene is dominated by ambiguous, repeating structures and presumably data augmentation helps to resolve some of this ambiguity.
On 12Scenes, we observe an improvement for RGB-only re-localization (+8.6\%), while the other results on this particular dataset are saturated also without data augmentation.
For the Cambridge Landmarks dataset, we found no significant advantage in augmenting the training data.
Notably, DSAC* achieves state-of-the-art accuracy across settings also without the use of data augmentation, cf.~Fig.~\ref{fig:exp:scenes_bars}.

\subsection{Impact of the Receptive Field} 
\label{sec:exp:rf}

\begin{figure}[!t]
    \centering
    \includegraphics[width=1\linewidth]{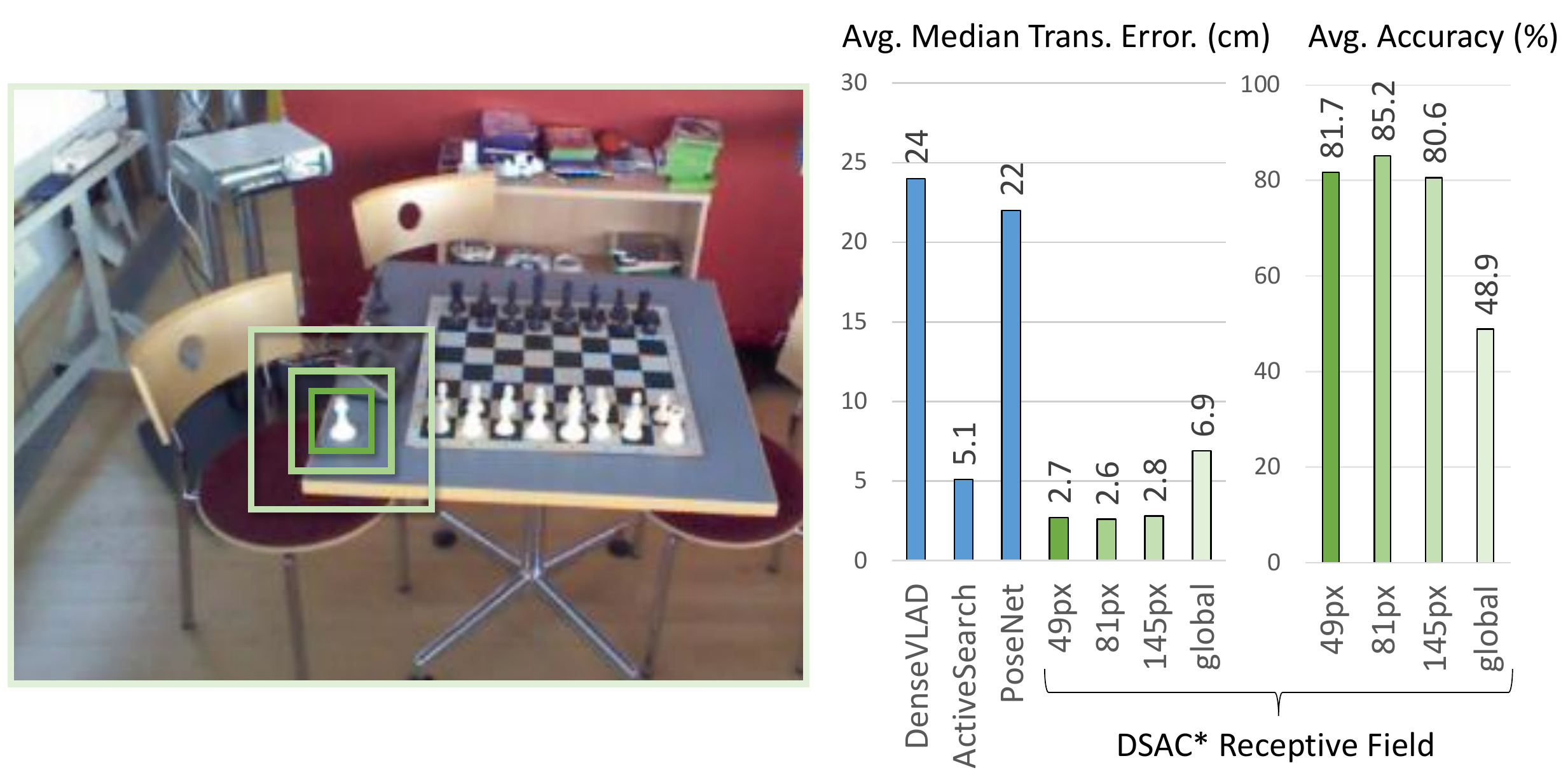}
    \vspace{-0.5cm}
    \caption{\textbf{Receptive Fields.} We study the impact of the receptive field for DSAC* by altering the underlying network architecture, see the main text for details. \textbf{Left.} We visualize the different receptive field sizes of DSAC* relative to a test image. \textbf{Right.} We report the median localization errors and percentages of re-localized frames for the 7Scenes dataset. \emph{Global} means a receptive field with the size of the whole image. DenseVLAD \cite{torii2015viewsynth} and PoseNet \cite{geometricloss} also utilize a global receptive field. ActiveSearch \cite{sattler2016efficient} utilizes a varying but limited receptive field.}
    \label{fig:exp:receptive}
\end{figure}

One important factor when designing an architecture for scene coordinate regression is the size of the receptive field.
That is, what image area is taken into account for predicting a single scene coordinate, comparable to the image patch size for sparse feature matching.
The architecture of DSAC* has a receptive field size of 81px.
By substituting individual 3x3 convolutions with 1x1 convolutions and vice versa (cf.~Fig.~\ref{fig:method:system}) we can increase and decrease the receptive field and study the change in accuracy.
The change of the convolution kernel affects also the total count of learnable parameters of the network.
To facilitate conclusions with regard to the receptive field alone, we scale the number of channels throughout the network to keep the number of free parameters constant.
We report results in Fig.~\ref{fig:exp:receptive}, comparing DSAC* with a receptive field of 81px (standard), 49px and 149px. 
We observe that the re-localization accuracy decreases with a large receptive field of 149px. 
While a larger receptive field incorporates more image context for predicting a scene coordinate, is also leads to generalization problems, even when using data augmentation during training.
View point changes between training and test set have a higher impact for larger receptive fields.
Making the receptive field smaller, with 49px, also decreases accuracy slightly.
The effect of having less image context is counteracted by better generalization \wrt view point changes.
For a more extreme argument in favor of architectures with limited receptive field, we conduct an experiment with an encoder-decoder architecture.
Such an architecture encodes the whole image into a global descriptor, and de-convolves it to a full resolution scene coordinate prediction.
The receptive field of such an architecture is the whole image, and we ensure again to keep the number of learnable parameters identical.
As depicted in Fig.~\ref{fig:exp:receptive} a scene coordinate network with global receptive field achieves a disappointing re-localization accuracy.
This indicates, that the receptive field might be another issue connected with the low accuracy of absolute pose regression methods, orthogonal to the explanations given by Sattler \etal in their study of these methods \cite{sattler2019limits}.

\subsection{Impact of End-to-End Training} 
\label{sec:exp:end2end}

\begin{figure*}[!t]
    \centering
    \includegraphics[width=1\linewidth]{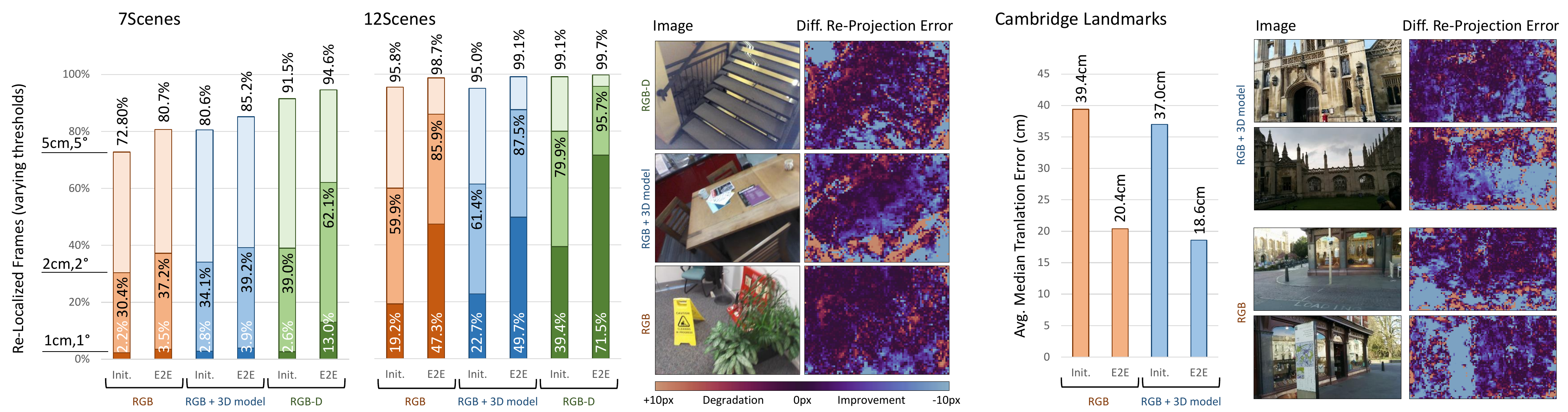}
    \vspace{-0.5cm}
    \caption{\textbf{Effect of End-to-End Training. Left:} We give the average percentage of correctly localized frames, on 7Scenes\cite{shotton13scorf} and 12Scenes\cite{valentin2016learning}, before and after end-to-end training, denoted as \emph{Init.} and \emph{E2E}, respectively. We break down accuracy corresponding to pose error thresholds of 5cm/5$^\circ$, 2cm/2$^\circ$ as well as 1cm/1$^\circ$. Colors in the bar chart indicate different training setups as also specified at the bottom. Furthermore, we visualize the \emph{difference} in re-projection error before and after end-to-end training for three training frames of 7Scenes \cite{shotton13scorf}. Blue areas indicate that the re-projection error decreased due to end-to-end training, red areas indicate that the error increased. \textbf{Right:} We show the median translation error, averaged over five scenes in Cambridge Landmarks \cite{kendall2015convolutional}, before and after end-to-end training, as well as a visualization of the change in re-projection error on training frames.}
    \label{fig:exp:e2e2}
\end{figure*}

We report report results before and after training our system in an end-to-end fashion in Fig.~\ref{fig:exp:e2e2}.
For the indoor datasets 7Scenes and 12Scenes we report accuracy using different threshold of 5cm5$^\circ$, 2cm2$^\circ$ and 1cm1$^\circ$.
While the impact of end-to-end training for a coarse threshold is small, there are significant differences for the finer acceptance thresholds.
End-to-end training increases the precision of successful pose estimates, but it does not necessarily decrease the failure rate. 
We see similar effects in outdoor re-localization for the Cambridge dataset where the pose precision, expressed by the median pose error decreases by ca.~30\%.
We also provide a qualitative comparison of scene coordinate prediction before and after end-to-end training. 
Particularly, we visualize areas of training images where the re-projection error increased or decreased due to end-to-end training.
The system learns to focus on certain reliable structures. 
In general, we observe a tendency of the system to increase the scene coordinate quality for close objects.
Presumably such objects are more helpful than distant structures for estimating the camera pose precisely.

\subsection{Learned 3D Geometry}
\label{sec:exp:geometry}

\begin{figure*}[!t]
    \centering
    \includegraphics[width=1\linewidth]{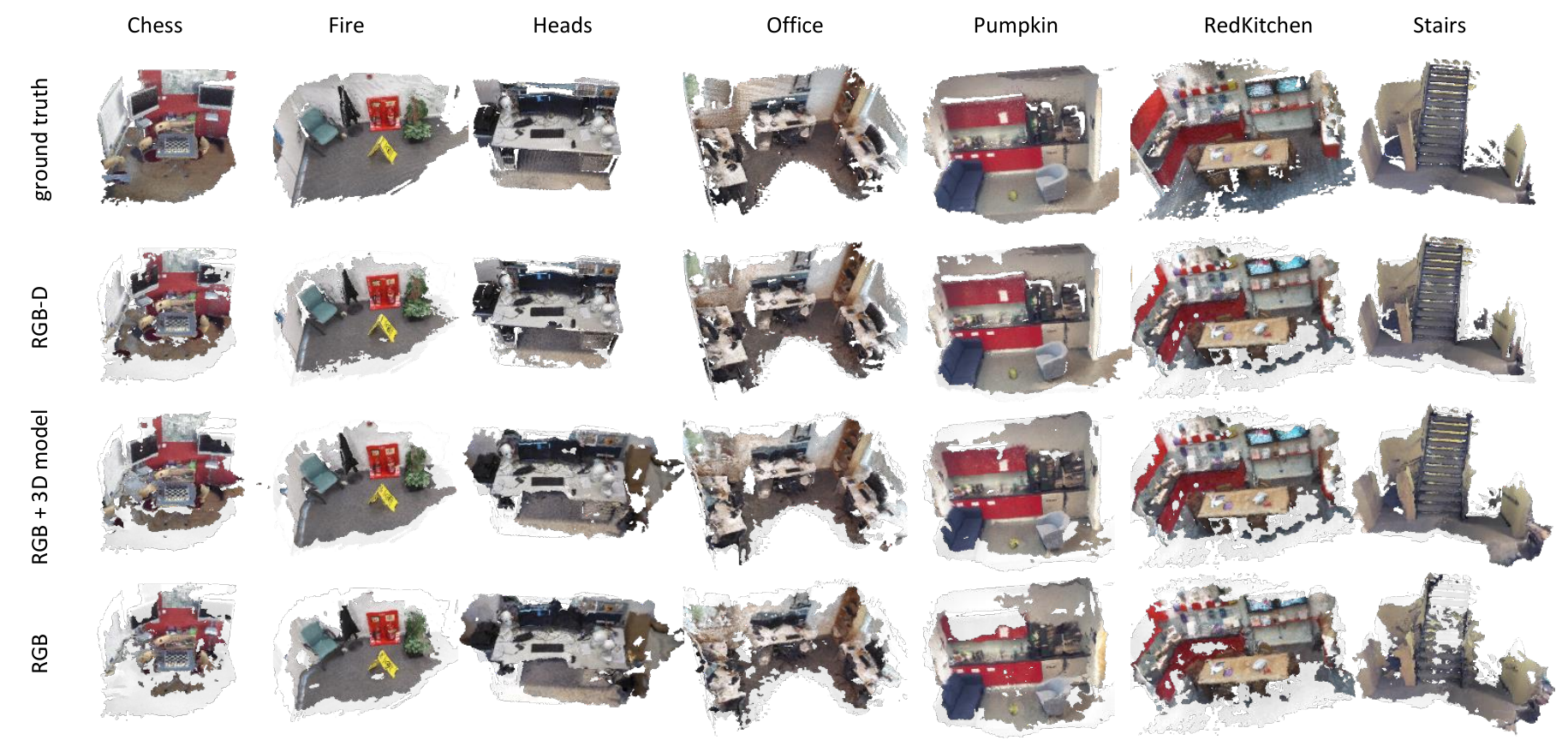}
    \vspace{-0.5cm}
    \caption{\textbf{Learned Indoor Geometries.} We visualize the 3D scene geometry learned by the scene coordinate regression network for all scenes of the 7Scenes \cite{shotton13scorf} dataset. See the main text for details on how we generated these models. Each row represents a different training setup. In particular, the last row, \emph{RGB}, shows geometry discovered by the network automatically given only RGB images and ground truth poses. For a more informative visualization, we always show the ground truth model as a faint backdrop.}
    \label{fig:exp:7scenes_models}
\end{figure*}

\begin{figure*}[!t]
    \centering
    \includegraphics[width=1\linewidth]{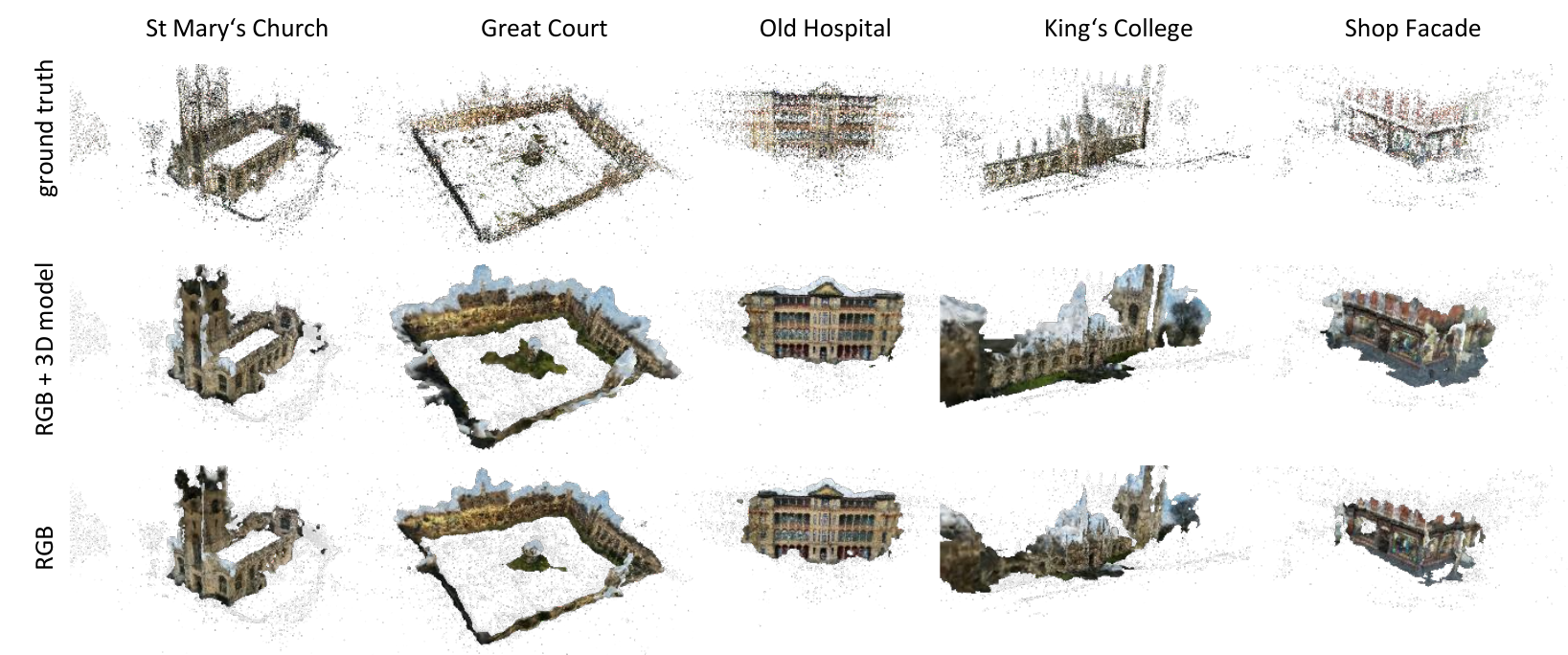}
    \caption{\textbf{Learned Outdoor Geometries.} We visualize the 3D scene geometry learned by the scene coordinate regression network for scenes of the Cambridge Landmarks \cite{kendall2015convolutional} dataset. See the main text for details on how we generated these models. Each row represents a different training setup. In particular, the last row, \emph{RGB}, shows geometry discovered by the network automatically given only RGB images and ground truth poses. For a more informative visualization, we always show the ground truth model as a faint backdrop. Note that the ground truth models of this dataset are sparse point clouds created by structure-from-motion tools.}
    \label{fig:method:cambridge_models}
\end{figure*}

Scene coordinate regression methods utilize a learnable function to implicitly encode the map of an environment. 
We can generate an explicit map representation of the geometry encoded in a network.
More precisely, we iterate over all training images, predicting scene coordinates to generate one point cloud of the scene. 
We can recover the color of each 3D point by reading out the associated color at the pixel position of the training image for which the scene coordinate was predicted.
Such a point cloud will in general feature many outlier points that hinder visualization.
Therefore, we generate a mesh representation using Poisson surface reconstruction \cite{kazhdan06poisson}.
We show the recovered 3D models in Fig.~\ref{fig:exp:7scenes_models} for 7Scenes and in Fig.~\ref{fig:method:cambridge_models} for Cambridge.
Interestingly, our approach learns the complex 3D geometry of a scene, even when training solely from RGB images and ground truth poses.
Furthermore, we are able to recover a dense scene representation, even when training with sparse 3D models for the Cambridge dataset. 

\subsection{Scene Compression Properties}
\label{sec:exp:compression}

Since scene coordinate regression methods encode the scene geometry within a neural network of fixed capacity, they represent a natural framework for scene compression. 
In Table \ref{tab:exp:compression}, we compare the memory demand and accuracy of several learning-based as well as classical re-localization methods.
The Cambridge Landmarks \cite{kendall2015convolutional} dataset is particularly interesting for this comparison, as it features scenes of varying sizes.
With a memory footprint of 28MB, DSAC* achieves highest average re-localization accuracy.
The retrieval-based DenseVLAD \cite{torii2015viewsynth} as well as the feature-based hybrid compression schema of Camposeco \etal \cite{compression2019cvpr} demand only very little memory but also suffer from low re-localization accuracy. 
To analyze the scene compression properties of DSAC* further, we re-train our pipeline with a significantly leaner network architecture, called \emph{DSAC* Tiny}. 
We clamp the number of channels per layer to 128 (cf.~Fig.~\ref{fig:method:system}) which results in a memory footprint of 4MB per scene. 
For this analysis, we train using the 3D scene model but without training data augmentation.
We found data augmentation to deteriorate results for such a low-capacity network.
While \emph{DSAC* Tiny} has a memory demand in the same magnitude as the hybrid compression schema of \cite{compression2019cvpr}, it achieves significantly higher accuracy.
We find that the loss in accuracy compared to the full 28MB model grows with the scene size and complexity, see \eg the results for \emph{St Mary's Church}.
For smaller scenes, such as \emph{Shop Facade}, the loss in accuracy is negligible. 
We trained \emph{DSAC* Tiny} also for the 7Scenes and 12Scenes datasets, and report an average re-localization accuracy of 73.6\% and 98.1\%, respectively. 
Therefore, \emph{DSAC* Tiny} is among the top-performing methods for indoor re-localization despite the small memory demand.

\begin{table*}[]
\centering
\begin{tabular}{r|c|c|c|c|c|c|c|c||c|c}
\multicolumn{1}{l|}{} & \multicolumn{2}{c|}{King's College} & \multicolumn{2}{c|}{Old Hospital} & \multicolumn{2}{c|}{Shop Facade} & \multicolumn{2}{c||}{St Mary's Church} & \multicolumn{2}{c}{Average} \\ \cline{2-11} 
\multicolumn{1}{l|}{} & \multicolumn{1}{c|}{\begin{tabular}[c]{@{}c@{}}MB\\ used\end{tabular}} & \multicolumn{1}{c|}{\begin{tabular}[c]{@{}c@{}}Median\\ error (m)\end{tabular}} & \multicolumn{1}{c|}{\begin{tabular}[c]{@{}c@{}}MB\\ used\end{tabular}} & \multicolumn{1}{c|}{\begin{tabular}[c]{@{}c@{}}Median\\ error (m)\end{tabular}} & \multicolumn{1}{c|}{\begin{tabular}[c]{@{}c@{}}MB\\ used\end{tabular}} & \multicolumn{1}{c|}{\begin{tabular}[c]{@{}c@{}}Median\\ error (m)\end{tabular}} & \multicolumn{1}{c|}{\begin{tabular}[c]{@{}c@{}}MB\\ used\end{tabular}} & \multicolumn{1}{c||}{\begin{tabular}[c]{@{}c@{}}Median\\ error (m)\end{tabular}} & \multicolumn{1}{c|}{\begin{tabular}[c]{@{}c@{}}MB\\ used\end{tabular}} & \multicolumn{1}{c}{\begin{tabular}[c]{@{}c@{}}Median\\ error (m)\end{tabular}} \\ \hline
Active Search \cite{sattler2016efficient} & 275MB & 0.57 & 140MB & 0.52 & 39MB & 0.12 & 359MB & 0.22 & 203MB & 0.36 \\
HybridCompression \cite{compression2019cvpr} & 1.0MB & 0.81 & 0.6MB & 0.75 & 0.2MB & 0.19 & 1.3MB & 0.50 & 0.8MB & 0.56 \\
DenseVLAD \cite{torii2015viewsynth} & 10MB & 2.8 & 14MB & 4.0 & 3.6MB & 1.1 & 23MB & 2.3 & 13MB & 2.6 \\
PoseNet17 \cite{geometricloss} & 50MB & 0.88 & 50MB & 3.2 & 50MB & 0.88 & 50MB & 1.6 & 50MB & 1.6 \\ \hline
DSAC++ \cite{brachmann2018lessmore} & 104MB & 0.18 & 104MB & 0.20 & 104MB & 0.06 & 104MB & 0.13 & 104MB & 0.14 \\
DSAC* & 28MB & 0.15 & 28MB & 0.21 & 28MB & 0.05 & 28MB & 0.13 & 28MB & 0.13 \\
DSAC* (Tiny) & 4MB & 0.19 & 4MB & 0.23 & 4MB & 0.07 & 4MB & 0.39 & 4MB & 0.22 \\ \hline
\end{tabular}
\caption{\textbf{Scene Compression Analysis.} We compare memory demand and re-localization accuracy of several methods on the Cambridge Landmarks dataset \cite{kendall2015convolutional}. Information of competitors taken from \cite{compression2019cvpr}. DSAC* (Tiny) is a variant of our network architecture with a reduced number of channels per layer.}
\label{tab:exp:compression}
\end{table*}

\section{Conclusion}
\label{sec:conclusion}

We have presented DSAC*, a versatile pipeline for single image camera re-localization based on scene coordinate regression and differentiable RANSAC.
In this article, we have derived gradients for all steps of robust pose estimation, including PnP solvers.
The resulting system supports \mbox{RGB-D}-based as well as RGB-based camera re-localization, and can be trained with or without a 3D model of a scene.
Compared to previous iterations of the system, DSAC* trains faster, needs less memory and features low runtime.
Simultaneously, DSAC* achieves state-of-the-art accuracy on various dataset, indoor and outdoor, and in various settings.
We made the code of DSAC* publicly available, and hope that is serves as a credible baseline in re-localization research.


%


\ifCLASSOPTIONcompsoc
  \section*{Acknowledgments}
\else
  \section*{Acknowledgment}
\fi

The authors would like to thank Dehui Lin for implementing an efficient version of the differentiable Kabsch pose solver within the scope of his Master thesis. 

This work was supported by the DFG grant COVMAP: Intelligente Karten mittels gemeinsamer GPS- und Videodatenanalyse (RO 4804/2-1 and RO 2497/12-2). This project has received funding from the European Research Council (ERC) under the European Unions Horizon 2020 programme (grant No. 647769). 

The computations were performed on an HPC Cluster at the Center for Information Services and High Performance Computing (ZIH) at TU Dresden.

\ifCLASSOPTIONcaptionsoff
  \newpage
\fi



\bibliographystyle{IEEEtran}
\bibliography{IEEEabrv,dsacstar}
%



%






\end{document}